\documentclass{article}

\usepackage{microtype}
\usepackage{graphicx}
\usepackage{subfigure}
\usepackage{booktabs} 
\usepackage{mathrsfs}

\usepackage{thm-restate}
\usepackage{mathtools}
\usepackage{nicefrac}
\usepackage{xcolor}
\usepackage{amsmath,amssymb,amsthm}

\theoremstyle{definition}
\newtheorem{theorem}{Theorem}

\newtheorem{lemma}[theorem]{Lemma}

\newtheorem{assumption}[theorem]{Assumption}

\usepackage{enumitem}
\setlist[enumerate]{leftmargin=0.5cm,topsep=0pt,itemsep=-2pt}
\setlist[itemize]{leftmargin=0.5cm,topsep=0pt,itemsep=-2pt}

\usepackage{hyperref}
\hypersetup{
    colorlinks=true,
    linkcolor=blue,
    citecolor=cyan,
}
\usepackage{mathrsfs}

\usepackage[accepted]{icml2021}

\icmltitlerunning{DoMo-AC: Doubly Multi-step Off-policy Actor-Critic Algorithm}

\begin{document}

\twocolumn[
\icmltitle{DoMo-AC: Doubly Multi-step Off-policy Actor-Critic Algorithm}



\icmlsetsymbol{equal}{*}

\begin{icmlauthorlist}
\icmlauthor{Yunhao Tang*}{dm}
\icmlauthor{Tadashi Kozuno*}{om}
\icmlauthor{Mark Rowland}{dm}
\icmlauthor{Anna Harutyunyan}{dm}\\
\icmlauthor{R\'emi Munos}{dm}
\icmlauthor{Bernardo \'Avila Pires}{dm}
\icmlauthor{Michal Valko}{dm}
\end{icmlauthorlist}

\icmlaffiliation{dm}{Google DeepMind}
\icmlaffiliation{om}{Omron Sinic X}

\icmlcorrespondingauthor{Yunhao Tang}{robintyh@deepmind.com}

\icmlkeywords{Machine Learning, ICML}

\vskip 0.3in
]



\printAffiliationsAndNotice{\icmlEqualContribution} 

\begin{abstract}
 Multi-step learning applies lookahead over multiple time steps and has proved valuable in policy evaluation settings. However, in the optimal control case, the impact of multi-step learning has been relatively limited despite a number of prior efforts. Fundamentally, this might be because multi-step policy improvements require operations that cannot be approximated by stochastic samples, hence hindering the widespread adoption of such methods in practice. To address such limitations, we introduce doubly multi-step off-policy VI (DoMo-VI), a novel oracle algorithm that combines multi-step policy improvements and policy evaluations. DoMo-VI enjoys guaranteed convergence speed-up to the optimal policy and is applicable in general off-policy learning settings. We then propose doubly multi-step off-policy actor-critic (DoMo-AC), a practical instantiation of the DoMo-VI algorithm. DoMo-AC introduces a bias-variance trade-off that ensures improved policy gradient estimates. When combined with the IMPALA architecture, DoMo-AC has showed improvements over the baseline algorithm on Atari-$57$ game benchmarks.
\end{abstract}

\section{Introduction}

Off-policy learning plays a central role in modern reinforcement learning (RL), where the algorithm learns from off-policy data such as exploratory actions, expert demonstrations and previous experiences. Off-policy learning consists of two critical components: \emph{off-policy evaluation}, where the aim is to approximate the value function of a target policy; and \emph{off-policy control}, where the aim is to approximate the optimal value function. Designing good evaluation and control algorithms are crucial to high-performing RL systems. %

In the meantime, multi-step learning has provided a robust and consistent improvement to policy evaluation. Unlike one-step bootstrapping methods such as TD($0$), multi-step learning bootstraps from predictions across multiple time steps along the trajectory, usually allowing for a much faster propagation of reward information across time. Empirically, this often helps the algorithm converge faster to the target value. In off-policy learning, notable examples 
include the Retrace and V-trace algorithms \citep{munos2016safe,espeholt2018impala}, which reduce to the celebrated 
TD($\lambda$) algorithm in the on-policy case \citep{sutton1998}.

In the control case, the most common approach is to find an improved policy by being greedy with respect to the current value function \citep{sutton1998}. The greedy improvement effectively looks ahead for a single time step, and intuitively should also benefit from multi-step learning as TD($0$). On the theory front, prior work has extended the one-step greedy improvement to the multi-step case \citep{efroni2018multiple,tomar2020multi}. However, a fundamental challenge is that since multi-step control consists of solving a optimal control problem in the inner loop \citep{efroni2018multiple}, it is not straightforward to combine such an approach with sample-based learning and incremental learning. 
As a result, this hinders the widespread adoption of multi-step learning, as it cannot be directly applied to policy improvement and optimal control. In this work, we aim to address the key question: how to make multi-step off-policy learning practical and theoretically sound for the control case? To this end, we make a few theoretical and practical contributions.

\paragraph{Doubly multi-step off-policy value iteration (DoMo-VI).} We introduce DoMo-VI, a multi-step learning algorithm consisting of multi-step policy evaluation and multi-step improvement (hence the name \emph{doubly}, Section~\ref{sec:vi}). DoMo-VI is compatible with using off-policy data, provably converges to the optimal value function with accelerated convergence rate, thanks to the application of multi-step learning to both the policy evaluation and improvement steps. To our knowledge, this is the first set of theoretical results on how multi-step control speeds up convergence in the off-policy setting.

\paragraph{Doubly multi-step off-policy actor-critic (DoMo-AC).} We introduce the DoMo-AC algorithm as a practical instantiation of DoMo-VI (Section~\ref{sec:ac}). The algorithm is designed to allow for a bias-variance trade-off in constructing policy gradient estimates from off-policy data. When implemented with the distributed learning architecture IMPALA, \citep{espeholt2018impala}, DoMo-AC achieves stable performance improvements over baseline methods. This provides evidence on multi-step control in large-scale settings.

\section{Background}
Consider a Markov decision process (MDP) represented as the tuple $\left(\mathcal{X},\mathcal{A},P_R,P,\gamma\right)$ where $\mathcal{X}$ is a finite state space, $\mathcal{A}$ the finite action space, $P_R:\mathcal{X}\times\mathcal{A}\rightarrow \mathcal{P}(\mathbb{R})$ the reward kernel, $P:\mathcal{X}\times\mathcal{A}\rightarrow\mathcal{P}(\mathcal{X})$ the transition kernel and $\gamma\in [0,1)$ the discount factor. 
For policy evaluation, the aim is to compute a value function $V^\pi(x)\coloneqq\mathbb{E}_\pi\left[\sum_{t=0}^\infty \gamma^tR_t\;\middle|\;X_0=x\right]$ for a target policy $\pi$; for optimal control, the aim is to find the optimal policy $\pi^\ast = \arg\max_{\pi \in \Pi} V^\pi$ from the set of all Markovian policies $\Pi$  \citep{puterman1990markov}.
\paragraph{Notation.} For careful readers, we provide a more precise definition of $\arg\max_{\pi \in \Pi} V^\pi$. Since $\mathcal{X}$ is finite, $V^\pi$ can be regarded as a $|\mathcal{X}|$-dimensional vector. We equip $\mathbb{R}^{|\mathcal{X}|}$ with the partial ordering induced by the non-negative orthant $[0, \infty)^{|\mathcal{X}|}$ as in \citet{boyd2004convex}. This ensures the maximization is well defined.

\subsection{Off-policy evaluation}

In off-policy evaluation, the aim is to compute approximations to a target value function $V^\pi$ given off-policy data 
 generated under a behavior policy $\mu:\mathcal{X}\rightarrow\mathcal{P}(\mathcal{A})$, which generally differs from the target policy $\pi$. 
As a standard assumption, we require the behavior policy $\mu$ to have full support over the action space: $\forall (x,a)\in\mathcal{X}\times\mathcal{A},\, \mu(a|x)>0$.

One general approach to off-policy evaluation is importance sampling (IS) \citep{precup2000eligibility,precup2001off}. 
Define step-wise IS ratio $\rho_t\coloneqq \pi(A_t|X_t)/\mu(A_t|X_t)$ and the trace coefficient $c_t=\min(\bar{c},\rho_t)$ with threshold $\bar{c}\geq 0$. Let $c_{0:t}\coloneqq c_0...c_t$ be the product of traces. The V-trace operator is defined as
\begin{align}
     \mathcal{R}_{\bar{c}}^{\pi,\mu}V(x) \coloneqq V(x) + \mathbb{E}_\mu\left[\sum_{t=0}^\infty \gamma^t c_{0:t-1} \rho_t\delta_t \right],
    \label{eq:V-trace}
\end{align}
with TD error $\delta_t\coloneqq R_t + \gamma V(X_{t+1})-V(X_t)$. The operator $\mathcal{R}_{\bar{c}}^{\pi,\mu}$ is $\eta$-contractive with some $\eta\in[0,\gamma]$ and has $V^\pi$ as the unique fixed point.  The threshold $\bar{c}$ determines the effective lookahead horizon for the operator. At one extreme $\bar{c}=0$, V-trace looks ahead for a single time step and reduces to the Bellman operator $\mathcal{T}^\pi V(x)\coloneqq \mathbb{E}_\pi\left[R_0+\gamma V(X_1) |\; X_0=x \right]$, for which $\eta=\gamma$, and the contraction is slow. At another extreme $\bar{c}=\infty$, V-trace looks ahead until the end of the trajectory and reduces to the IS evaluation in expectation $\mathcal{R}_{\bar{c}}^{\pi,\mu}V(x)=V^\pi(x)$. In this case, the contraction is fast $\eta=0$ but stochastic approximations to the V-trace target can have high variance. In practice, it is common to apply $\bar{c}=1$ to achieve a better contraction-variance trade-off \citep{espeholt2018impala,munos2016safe}.

\subsection{Optimal control by value iteration}
Value iteration (VI) is one primary approach for finding the optimal policy $\pi^\ast$. VI is a recursion on the
policy and value function pair $(\pi_{i+1}, V_i)_{i=0}^\infty$, which include a policy improvement step and a policy evaluation step \citep{puterman1990markov}:
\begin{align}
    \pi_{i+1}(\cdot|x) &= \arg\max_{\pi \in \Pi}  \mathcal{T}^\pi V_i(x) \tag{policy improvement},\\  V_{i+1} &= \mathcal{T}^{\pi_{i+1}}V_i.\tag{policy evaluation}
\end{align}

In the policy improvement step, $\pi_{i+1}$ extracts the greedy policy at state $x$ based on the one-step lookahead objective 
$
    \arg\max_a \mathbb{E}\left[R_0+\gamma V(X_1)\;\middle|\;X_0=x,A_0=a\right]
$.
In the policy evaluation step, $V_{i+1}=\mathcal{T}^{\pi_{i+1}}V_i\approx V^{\pi_{i+1}}$ approximates the value function of the improved policy $\pi_{i+1}$.

A potential drawback of VI is that it carries out only \emph{shallow} policy improvement and policy evaluation. The policy improvement step looks ahead for a single time step $R_0+\gamma V(X_1)$, which may result in slow improvement
\citep{efroni2018multiple,tomar2020multi}. For policy evaluation, one single application of the Bellman operator $\mathcal{T}^{\pi_{i+1}}$ might not be accurate enough due to slow contraction of the operator.

\section{Doubly multi-step off-policy VI (DoMo-VI)}
\label{sec:vi}

To alleviate the shallow policy improvement and evaluation of VI, we propose the following 
DoMo-VI recursions
\begin{align}
    \pi_{i+1}(\cdot|x) &= \arg\max_{\pi \in \Pi}  \mathcal{R}_{\bar{c}}^{\pi,\mu} V_i(x) \nonumber,\\  V_{i+1} &= \mathcal{R}_{\bar{c}}^{\pi_{i+1},\mu} V_i.\label{eq:offpolicy-multistep-vi}
\end{align}
By setting $\bar{c}=0$ such that V-trace reduces to the one-step Bellman operator, DoMo-VI reduces to VI. When $\bar{c}>0$, the improvement objective $\mathcal{R}_{\bar{c}}^{\pi,\mu} V_i$ effectively looks ahead multiple steps starting from $x$, resulting in a stronger improvement when the maximization problem can be solved exactly.
Indeed, at the extreme when $\bar{c}=\infty$, the improvement objective becomes the value function $\mathcal{R}_{\bar{c}}^{\pi,\mu} V_i = \arg\max_{\pi \in \Pi} V^\pi$ and the improvement step returns the optimal policy $\pi^\ast$. 

One subtle technical question is whether the above maximization is well defined, i.e., whether there exists a single Markov policy $\pi$ which achieves the maximum. Fortunately, this is indeed the case.

\begin{lemma}\label{lemma:stationary} (\textbf{Optimal Markov policy}) For any real-valued function $V$ over $\mathcal{X}$, a scalar $\bar c$, and a behavior policy $\mu$, there exists a Markov policy $\pi$ such that $\pi = \arg\max_p  \mathcal{R}_{\bar{c}}^{p, \mu} V$.
\end{lemma}
Lemma~\ref{lemma:stationary} implies that we can obtain a single Markov policy that maximizes the improvement objective $\mathcal{R}_{\bar{c}}^{\pi,\mu}$ simultaneously across all states $x$. In practice, this means it is feasible to find the optimally improved policy according to the improvement objective $\mathcal{R}_{\bar{c}}^{\pi,\mu} V_i$. Such an improvement step can be carried out by a policy optimization subroutine. In general, when computing the exact optimal solution is too expensive, the optimization subroutine can be replaced by incremental updates, such as the policy gradient algorithm. We will discuss such a practical approach in Section~\ref{sec:ac}.

\begin{table*}[t!]
    \centering
    \caption{A list of algorithms that can be decomposed into a policy improvement (PI) step and a policy evaluation (PE) step. The convergence rate measures how fast $V^{\pi_i}$ converges to the optimal value function $V^\ast$. Concretely, if an algorithm's performance is bounded as $\left\lVert V^{\pi_i}- V^\ast\right\rVert_\infty\leq \eta^i C$ for some constant $C$. Here, $\eta\in[0,1]$ is the convergence rate. The list of algorithms include 
    (1) multi-step PE, which closely relates to Q($\lambda$), Retrace and Peng's Q($\lambda$) in the control case \citep{harutyunyan2016q,munos2016safe,peng1994incremental,kozuno2021revisiting}; (2) multi-step PI, which relates to $\lambda$-VI in the on-policy case \citep{efroni2018multiple}; (3) one-step baseline VI, and (4) $\lambda$-policy iteration \citep{efroni2018multiple}, which requires a PE oracle. 
    \newline}
    \begin{sc}
    \begin{tabular}{l|c|c|c}\toprule[1.5pt]
        \bf Algorithm & \bf Policy improvement & \bf Policy evaluation & \bf Convergence rate \\\midrule
       DoMo-VI & $\pi_{i+1}(\cdot|x) = \arg\max_{\pi \in \Pi} \mathcal{R}_{\bar{c}}^{\pi,\mu}   V_i(x)$ & $V_{i+1} = \mathcal{R}_{\bar{c}}^{\pi_{i+1},\mu}   V_i$ & $\eta^\ast\in[0,\gamma]$
        \\
        Multi-step PE only & $\pi_{i+1}(\cdot|x) = \arg\max_{\pi \in \Pi} \mathcal{T}^\pi  V_i(x)$ & $V_{i+1} = \mathcal{R}_{\bar{c}}^{\pi_{i+1},\mu}   V_i$ & NA
        \\
        Multi-step PI only & $\pi_{i+1}(\cdot|x) = \arg\max_{\pi \in \Pi} \mathcal{R}_{\bar{c}}^{\pi,\mu}   V_i(x)$ & $V_{i+1} = \mathcal{T}^{\pi_{i+1}}  V_i$ & NA
        \\
        Value Iteration & $\pi_{i+1}(\cdot|x) = \arg\max_{\pi \in \Pi} \mathcal{T}^\pi  V_i(x)$ & $V_{i+1} = \mathcal{T}^{\pi_{i+1}}  V_i$ & $\gamma$
        \\
        $\lambda$-policy iteration & $\pi_{i+1}(\cdot|x) = \arg\max_{\pi \in \Pi} \mathcal{T}_\lambda^\pi V_i(x)$ & $V_{i+1} = V^{\pi_{i+1}}$ & $\frac{\gamma(1-\lambda)}{1-\gamma\lambda}$
        \\
        \bottomrule[1.46pt]
    \end{tabular}
    \end{sc}
    \label{table:list of algorithms}
\end{table*}

\subsection{Convergence of DoMo-VI}

We now show that DoMo-VI converges to the optimal policy $\pi^\ast$ at an accelerated convergence rate.
\begin{theorem}\label{theorem:optimal} (\textbf{Convergence rate to optimality})
Assume that expected rewards take values in $[-\bar{R}, \bar{R}]$, and $V_0$ is bounded by $1/(1-\gamma)$. Then, there exist a scalar $\eta^\ast\in[0,\gamma]$ and a sequence of scalars $(\eta_j)_{j=1}^{\infty}$ in $[0,\gamma]$ such that DoMo-VI (Eqn~\eqref{eq:offpolicy-multistep-vi}) generates a sequence of Markov policies $(\pi_i)_{i=1}^{\infty}$ with value functions satisfying the following guarantee:
\begin{align*}
    \left\lVert V^{\pi_{i+1}}-V^\ast \right\rVert_\infty \leq \max \left\{\left( \eta^\ast \right)^i, \prod_{j=1}^i \eta_j \right\} \frac{4\bar{R}}{(1-\gamma)^2}\,.
\end{align*}
\end{theorem}

The above result shows that DoMo-VI generates policy sequence $\pi_i$ whose performance $V^{\pi_i}$ converges to the optimal performance $V^\ast$. The convergence rate depends on $\eta^\ast$ and $(\eta_j)_{j=1}^{\infty}$. It is useful to examine the explicit form of the contraction rate \citep{espeholt2018impala}. Let us consider only $\eta^\ast$ for simplicity. It holds that
\begin{align*}
\eta^\ast&=\mathbb{E}_\mu\left[\sum_{t=1}^\infty \gamma^t c_{0:t-2} \left(1-c_{t-1}\right)\right] \\ 
&=\gamma \left(1-\mathbb{E}_\mu[c_0]\right) + \gamma^2 \left(\mathbb{E}_\mu[c_0]-\mathbb{E}_\mu[c_0c_1]\right) + ...
\end{align*}
When $\bar{c}=0$, the above result recovers the convergence rate of one-step VI, which is $\gamma^i$. When $\bar{c}$ is large and there is little truncation on the IS ratio $\pi^\ast(a|x)/\mu(a|x)$, the contraction rate is small $\eta^\ast\approx 0$ and the convergence to optimality takes place in one iteration. For intermediate values of $\bar{c}$, since $\eta^\ast \leq  \gamma$ and $\eta_i\leq \gamma$, we expect a speed up to the convergence rate of VI. 

The accelerated convergence rate comes at a cost, as much of the computational complexity is hidden under the policy improvement step $\arg\max_{\pi \in \Pi} \mathcal{R}_{\bar{c}}^{\pi,\mu}V$. Since $\bar{c}$ determines the lookahead horizon of the V-trace operator, it also determines how difficult to solve the inner loop optimization problem exactly. When $\bar{c}=\infty$ and $\eta^\ast=0$, the policy improvement step effectively reduces to solving the control problem itself $\arg\max_{\pi \in \Pi} V^\pi$. 
In practice, $\bar{c}$ mediates a trade-off between the inner loop complexity of multi-step policy improvement and outer loop convergence rate. As we will show empirically, approximately optimizing the policy improvement objective suffices to speed up convergence (Section~\ref{sec:exp})

\begin{figure}[t]
    \centering
    \includegraphics[keepaspectratio,width=.4\textwidth]{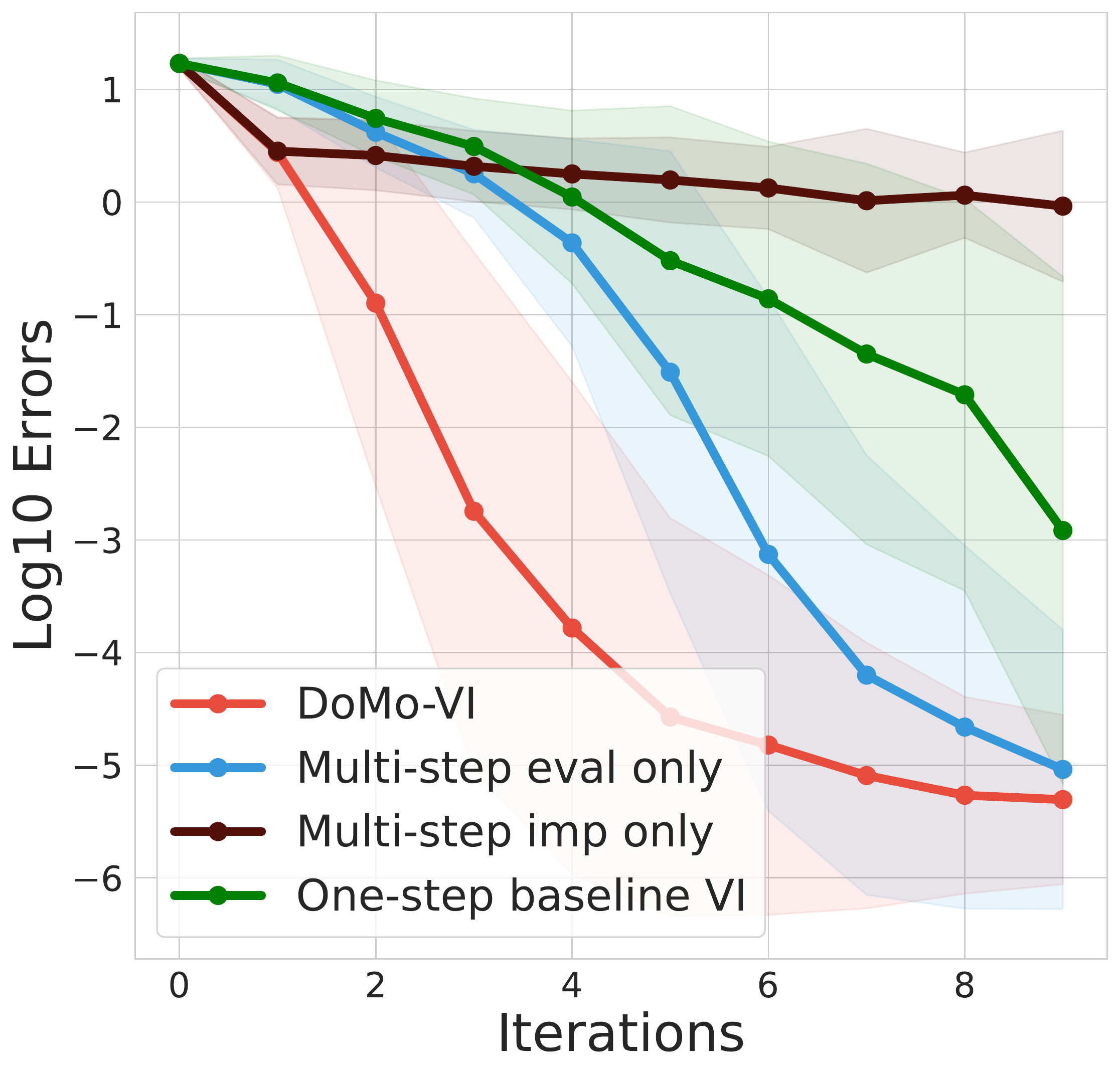}
    \caption{Comparing DoMo-VI with multi-step policy evaluation only (similar to \citep{espeholt2018impala,munos2016safe}), multi-step policy optimization (similar to \citep{efroni2018multiple}) and one-step baseline VI. The $y$-axis shows the value error $\left\lVert V^{\pi_i} - V^\ast \right\rVert$ on tabular MDPs. DoMo-VI combines the strengths of both multi-step policy evaluation and optimization, and achieves the fastest convergence rate among all baselines. Results are averaged across $100$ runs on tabular MDPs. See Appendix~\ref{appendix:exp} for details.}
    \label{fig:rate}
\end{figure}

\subsection{Understanding DoMo-VI}

Next we discuss algorithms that interpolate VI and DoMo-VI. This helps decompose the performance improvement of DoMo-VI, and sheds light on the design choice of the algorithm. In Table 1, we make a list of algorithms that interpolate VI and DoMo-VI, as well as a number of highly related algorithms in prior literature.

\paragraph{Multi-step policy evaluation.}
Starting with VI, let us first seek to remedy shallow policy evaluation in VI. We can replace the one-step operator $\mathcal{T}^\pi$ by the V-trace operator $\mathcal{R}_{\bar{c}}^{\pi,\mu}$ for policy evaluation, resulting in the following recursion of \emph{multi-step policy evaluation},
\begin{align}
    \pi_{i+1}(\cdot|x) = \arg\max_{\pi \in \Pi}  \mathcal{T}^{\pi} V_i(x) , \ \  V_{i+1} =  \mathcal{R}_{\bar{c}}^{\pi_{i+1},\mu} V_i.\tag{multi-step policy evaluation}
\end{align}
Such a recursion bears close connections to algorithms such as Q($\lambda$), Retrace and Peng's Q($\lambda$) in the control case \citep{harutyunyan2016q,munos2016safe,kozuno2021revisiting}. The aim of such algorithm is to improve the convergence speed of the policy evaluation step. In the extreme when $\bar{c}=\infty$, the evaluation is exact $V_{i+1}=V^{\pi_{i+1}}$ and the above recursion is equivalent to policy iteration (PI), which empirically at a much faster rate than VI to the optimal policy \citep{puterman1990markov,scherrer2012approximate}. 

\paragraph{Multi-step policy improvement.}
Next, we can replace the one-step operator $\mathcal{T}^\pi$ by the V-trace operator $\mathcal{R}_{\bar{c}}^{\pi,\mu}$ for policy improvement. This leads to the following recursion of \emph{multi-step policy improvement},
\begin{align}
    \pi_{i+1}(\cdot|x) = \arg\max_{\pi \in \Pi} \mathcal{R}_{\bar{c}}^{\pi,\mu}   V_i(x) ,\ \   V_{i+1} =  \mathcal{T}^{\pi_{i+1}}  V_i. \tag{multi-step policy improvement}
\end{align}
In the on-policy case $\pi=\mu$ and $c_t=\lambda\in[0,1]$, the V-trace operator is equivalent to the on-policy TD($\lambda$) operator $\mathcal{R}_{\bar{c}}^{\pi,\mu} =\mathcal{T}_\lambda^\pi$. As a result, the above recursion recovers the multi-step greedy algorithm $\lambda$-VI  proposed in \citep{efroni2018multiple,tomar2020multi}. 

Finally, DoMo-VI can be understood as combining the strengths of both multi-step policy evaluation and multi-step policy improvement. In a tabular setting, we make a comparison between DoMo-VI and multiple algorithmic variants discussed above (see Figure~\ref{fig:rate}). Multi-step evaluation takes up most performance improvements from baseline VI, speeding up the convergence of $V^{\pi_i}$ to $V^\ast$. Perhaps surprisingly, multi-step policy optimization provides an initial speed up, but ultimately falls short even compared to the baseline.

\section{Doubly multi-step off-policy actor-critic (DoMo-AC)}
\label{sec:ac}

Now, we present the core practical algorithm DoMo-AC. Starting with DoMo-VI in Eqn~\eqref{eq:offpolicy-multistep-vi}, note that in general it is computationally expensive to exactly solve the maximization problem  that defines the policy improvement step $\arg\max_{\pi \in \Pi} \mathcal{R}_{\bar{c}}^{\pi,\mu}V_i(x)$. Instead, it is more tractable to take a single gradient step from the current policy iterate. When the policy is parameterized $\pi_\theta$, the update in the parameter space at state $x$ is
\begin{align}
    \theta_{i+1} &= \theta_i +  \beta \nabla_{\theta_i}  \mathcal{R}_{\bar{c}}^{\pi_i,\mu} V_i(x),
\end{align}
where $\beta>0$ is the learning rate.
Note that going from $\theta_i$ to $\theta_{i+1}$, the policy locally increases the policy improvement objective $\mathcal{R}_{\bar{c}}^{\pi_i,\mu} V_i(x)$. For general parameterization where $\theta\in\mathbb{R}^d$ is a vector in some $d$-dimensional Euclidean space, policies at different states share parameters. The policy update requires averaging gradient updates under a weighting distribution over state $x\sim b$. The combined recursion is hence
\begin{align*}
    \theta_{i+1} = \theta_i + \beta \mathbb{E}_{x\sim b}\left[\nabla_{\theta_i} \mathcal{R}_{\bar{c}}^{\pi_{\theta_i},\mu}V_i(x)\right],\ \ 
    V_{i+1}=\mathcal{R}_{\bar{c}}^{\pi_{\theta_i},\mu}V_i.
\end{align*}
We can interpret the above recursion as an an actor-critic algorithm, where the value function $V_i$ serves as the critic. Intriguingly, when $\bar{c}=0$, the policy update reduces to
\begin{align*}
    \theta_{i+1} = \theta_i + \beta\mathbb{E}\left[ \left(R_0+\gamma V(x')\right)\nabla_{\theta_i} \log \pi_{\theta_i}(a|x)\right],
\end{align*}
where the expecation is under $x\sim b,a\sim \pi_{\theta_i}(\cdot|x), x'\sim P(\cdot|x,a)$. This bears close resemblance to practical policy gradient updates adopted in high-performing policy-based deep RL agents \citep{wang2016sample,mnih2016,schulman2017,espeholt2018impala}. 

To derive properties for the gradient update, we assume a smoothly differentiable parameterization of the policy.
\begin{assumption} (\textbf{Smooth policy}) \label{assume:smooth}
The policy $\pi_\theta(a|x)$ is differentiable with respect to $\theta$ and $\left\lVert\frac{\partial \pi_\theta(a|x)}{\partial \theta}\right\rVert_\infty\leq G$ for some constant $G\geq 0$ and for all $ (x,a)\in\mathcal{X}\times\mathcal{A}$.
\end{assumption}

\subsection{Approximation to  policy gradient update}
At the extreme when $\bar{c}=\infty$, $\mathcal{R}_{\bar{c}}^{\pi_{\theta_i},\mu}V_i(x)\approx V^{\pi_{\theta_i}}(x)$ and the policy update reduces to an exact policy gradient update averaged over state distribution $x\sim b$,
\begin{align*}
    \theta_{i+1} &= \theta_i+  \beta \mathbb{E}_{x\sim b}\left[\nabla_{\theta_i} V^{\pi_{\theta_i}}(x)\right].
\end{align*}
Such an update is potentially desirable because it locally improves the average value function objective $\mathbb{E}_{x\sim b}\left[V^{\pi_\theta}(x)\right]$. In general when $\bar{c}$ is finite, the update may not locally improve the value function objective since $\nabla_{\theta_i}\mathcal{R}_{\bar{c}}^{\pi_{\theta_i},\mu} V_i(x)$ differs from the policy gradient direction $\nabla_{\theta_i} V^{\pi_{\theta_i}}(x)$. To clarify the effect of $\bar{c}$ on how well $\nabla_{\theta_i}\mathcal{R}_{\bar{c}}^{\pi_{\theta_i},\mu} V_i(x)$ carries out local improvement, we characterize its difference from the exact policy gradient.
\begin{theorem}\label{theorem:bias} (\textbf{Approximating policy gradient})
Recall $\eta$ to be the contraction rate of the V-trace operator $\mathcal{R}_{\bar{c}}^{\pi_\theta,\mu}$. Let $\theta_j$ be any scalar component of parameter $\theta\in\mathbb{R}^d$ and recall $V \in \mathbb{R}^{\mathcal{X}}$ to be a value function vector. Then $\nabla_{\theta_j} V^{\pi_\theta}\in\mathbb{R}^{\mathcal{X}}$ is a policy gradient vector over state for parameter $\theta_j$. Assume $V=V^{\pi_\theta}$, then
\begin{align*}
    &\left\lVert \nabla_{\theta_j}\mathcal{R}_{\bar{c}}^{\pi_{\theta},\mu} V - \nabla_{\theta_j} V^{\pi_\theta} \right\rVert_\infty  \leq \eta  \left\lVert \nabla_{\theta_j} V^{\pi_\theta}\right\rVert_\infty .
\end{align*}
\end{theorem}
We offer some interpretations of the above result. Note that even if the value function is perfectly evaluated $V=V^{\pi_\theta}$, there is an irreducible error as characterized by the error bound. To see why, recall the exact policy gradient as
\begin{align*}
    \nabla_{\theta} V^{\pi_\theta}(x) = \mathbb{E}_{\pi_\theta}\left[\sum_{t=0}^\infty \gamma^t\sum_a Q^{\pi_\theta}(X_t,a)\nabla_\theta \pi_\theta(a|X_t)\right].
\end{align*}
Let $\bar{c}=0$ and $V=V^{\pi_\theta}$, the approximate gradient is 
\begin{align*}
    \nabla_{\theta}\mathcal{R}_{\bar{c}}^{\pi_{\theta},\mu} V(x) = \sum_a  Q^{\pi_\theta}(x,a) \nabla_\theta \pi_\theta(a|x),
\end{align*}
which corresponds to the term at $t=0$ of the exact policy gradient. hence, we can indeed interpret the truncation threshold $\bar{c}$ as determining the lookahead horizon when calculating the policy gradient estimates, which become more accurate when $\bar{c}$ increases. This effect is reflected by the contraction rate $\eta$ in the error bound. 

Though a large value of $\bar{c}$ decreases the bias of the gradient estimate against the true policy gradient, it can also lead to high variance in the stochastic gradient estimates. We will examine such a bias-variance trade-off numerically in Section~\ref{sec:exp}.

\subsection{Low-variance unbiased gradient estimate}

In general, it is challenging to compute the gradient update exactly. Instead, it is more computationally desirable to construct unbiased gradient estimate with stochastic samples.
To this end, we recall that since the V-trace back-up target $\mathcal{R}_{\bar{c}}^{\pi_{\theta},\mu}V$ can be approximated by off-policy stochastic estimates in an unbiased way, this naturally leads to an unbiased estimate to $\nabla_{\theta} \mathcal{R}_{\bar{c}}^{\pi_{\theta},\mu}V$.

\begin{theorem}\label{theorem:unbiased} (\textbf{Unbiased gradient estimate})
Assume trajectories $(X_t,A_t,R_t)_{t=0}^\infty\sim \mu$ reach a terminal state within $H<\infty$ steps almost surely. Let $X_0=x$ be the initial state, the unbiased V-trace back-up target estimate is
\begin{align}
    \widehat{\mathcal{R}_{\bar{c}}^{\pi_{\theta},\mu}}V(x) \coloneqq V(x) + \sum_{t=0}^\infty \gamma^t c_{0:t-1} \rho_t\delta_t.\label{eq:V-trace-estimate}
\end{align}
 Further, $ \widehat{\mathcal{R}_{\bar{c}}^{\pi_{\theta},\mu}}V(x)$ is differentiable and $ \nabla_\theta \widehat{\mathcal{R}_{\bar{c}}^{\pi_{\theta},\mu}}V(x)$ is an unbiased estimate to $\nabla_\theta \mathcal{R}_{\bar{c}}^{\pi_{\theta},\mu}V(x)$.
\end{theorem}

Intriguingly, the naive estimate based on Eqn~\eqref{eq:V-trace-estimate} turns out to have low variance. To see this, consider the special case when $\bar{c}=\infty$ and the trace coefficient is effectively the step-wise IS ratio $c_t=\rho_t$. In this case, the gradient estimate evaluates to
\begin{align}
    \nabla_\theta \widehat{\mathcal{R}_{\bar{c}}^{\pi_{\theta},\mu}}V(x) = \sum_{t=0}^\infty \gamma^t \rho_{0:t} \widehat{A}_t \nabla_\theta \log \pi_\theta(A_t|X_t),\label{eq:special-case}
\end{align}
where $\widehat{A}_t=R_t + \gamma \widehat{V}(X_{t+1}) - V(X_t)$ is the advantage estimate. Here, the built-in variance reduction technique is the subtraction of value function $V(X_t)$ as a baseline when computing advantage estimate $\widehat{A}_t$, which is most commonly used in policy gradient estimate \citep{sutton1999,weaver2013optimal}. Secondly, the value estimate $\widehat{V}(X_t)$ turns out to be the doubly-robust value function estimate \citep{jiang2016doubly,thomas2016data}, which writes recursively as
\begin{align*}
    \widehat{V}(X_{t}) = V(X_t) + \rho_t \left(R_t + \gamma \widehat{V}(X_{t+1}) - V(X_t) \right).
\end{align*}
The doubly-robust estimation technique has also been known to reduce variance in off-policy learning \citep{jiang2016doubly,thomas2016data}. 
For general values of the trace coefficient $c_t$, we should expect a similar variance reduction effect.

\begin{algorithm}[t]
\label{algo:ac}
\begin{algorithmic}
\STATE Policy parameter $\theta_0$, critic parameter $\phi_0$ and target parameter $\phi_0^-$.
\FOR{$i=0,1,2...$}
\STATE \textbf{Collect data.} Collect trajectories $(X_t,A_t,R_t)_{t=0}^{T-1}$ of length $T$ under behavior policy $\mu$.
\STATE \textbf{Actor update.} Update policy $\pi_{\theta_i}$ based on Eqn~\eqref{eq:policy-update}.
\STATE \textbf{Critic update.} Update critic $V_{\phi_i}$ based on Eqn~\eqref{eq:critic-update}. and update target network.
\ENDFOR
\STATE  Output the final policy.
\caption{Doubly multi-step off-policy actor-critic (DoMo-AC)}
\end{algorithmic}
\end{algorithm}

\subsection{Implementation with function approximation}

Finally, we spell out the algorithm with both a parameterized policy $\pi_\theta$ and a parameterized critic $V_\phi$. Given a trajectory $(X_t,A_t,R_t)_{t=0}^{T-1}$ of length $T$, sampled under the behavior policy $\mu$, the policy is updated via the DoMo-AC gradient estimate
\begin{align}
    \theta_{i+1} = \theta_i + \beta \frac{1}{T}\sum_{t=0}^{T-1} \nabla_{\theta_i} \widehat{\mathcal{R}_{\bar{c}}^{\pi_{\theta_i},\mu}}V_{\phi_i}(X_t). \label{eq:policy-update}
\end{align}
Meanwhile, the critic is updated using gradient descent on the least square loss function
\begin{align}
    \phi_{i+1} = \phi_i - \beta \frac{1}{T}\sum_{t=0}^{T-1} \nabla_{\phi_i}  \left(V_\text{target}(X_t) - V_{\phi_i}(X_t)\right)^2, \label{eq:critic-update}
\end{align}
where $V_\text{target}(X_t)=\widehat{\mathcal{R}_{\bar{c}}^{\pi_{\theta_{i+1}},\mu}}V_{\phi_i^-}(X_t)$ is the back-up target computed via the target network $\phi_i^-$.  The target network is slowly updated towards the main network $\phi_i^-=(1-\tau)\phi_i^- + \tau \phi_i$ \citep{lillicrap2015continuous}. In practical implementations, it is more common to carry out the above gradient updates simultaneously. See Algorithm 1 for full algorithm.

\section{Discussion}

We provide discussions on a few lines of related work and natural extensions of our current method.

\paragraph{$\lambda$-policy iteration ($\lambda$-PI).} Another important variant of multi-step policy improvement algorithm is $\lambda$-PI \citep{efroni2018multiple}, which in our notations can be expressed as 
\begin{align*}
    \pi_{i+1}(\cdot|x) = \arg\max_{\pi \in \Pi} \mathcal{T}_\lambda^\pi   V_i(x) ,\ \   V_{i+1} =  V^{\pi_{i+1}},
\end{align*}
where $\mathcal{T}_\lambda^\pi$ is the on-policy TD($\lambda$) operator.
This algorithm achieves a convergence rate of $\frac{\gamma(1-\lambda)}{1-\lambda\gamma}$ to the optimal value function, which significantly speeds up one-step VI when $\lambda$ is close to $1$. One primary bottleneck of $\lambda$-PI is that it requires a policy evaluation oracle, setting the value function estimate $V_{i+1}$ to be the exact value function $V^{\pi_{i+1}}$. Such a critic is in general not accessible in practice. DoMo-VI removes such a limitation and replaces the oracle by a multi-step evaluation operator $V_{i+1}=\mathcal{R}_{\bar{c}}^{\pi_{i+1},\mu}V_i$, which can be practically implemented. Another major difference between DoMo-VI and $\lambda$-PI is that the latter requires on-policy data when doing policy improvement. 

\paragraph{Off-policy corrections are important for multi-step policy improvement.}
DoMo-VI can be extended to evaluation operators $\mathcal{R}_{\bar{c}}^{\pi,\mu}$ beyond V-trace, such as the value function variant of Q($\lambda$) \citep{harutyunyan2016q}, where the trace coefficient $c_t=\lambda$. This closely resembles TD($\lambda$) with the main difference being that the data is off-policy. The tree-backup trace $c_t=\pi(A_t|X_t)$ can be understood as a special case of V-trace \citep{precup2001off} since $c_t\leq \rho_t$. A primary bottleneck of tree-backup is that it cuts traces quickly and is not efficient when near on-policy \citep{munos2016safe}.
Another alternative is the value function equivalent of Peng's Q($\lambda$) operator \citep{peng1994incremental}, which can be understood as geometrically weighted sum of $n$-step TD($n$) operators. Unlike V-trace and Q($\lambda$), which carry out off-policy corrections, Peng's Q($\lambda$) does not have the target value function as the fixed point. Nevertheless, Peng's Q($\lambda$) has displayed practical benefits over methods based on proper off-policy corrections, thanks to its significant improvement in the contraction rate (though to the biased fixed point) \citep{kozuno2021revisiting}.

However, we can verify that when $\mathcal{R}_{\bar{c}}^{\pi,\mu}$ is the Peng's Q($\lambda$) operator, 
$\arg\max_{\pi \in \Pi}\mathcal{R}_{\bar{c}}^{\pi,\mu}V(x) $ corresponds to the one-step greedy policy. This means uncorrected algorithms such as Peng's Q($\lambda$) cannot entail multi-step policy improvement.

\paragraph{Multiple applications of  evaluation operator.} We can consider a more general form of the DoMo-AC gradient update, by differenting through multiple applications of the evaluation operator
\begin{align*}
    \theta_{i+1} = \theta_i + \mathbb{E}_{x\sim b}\left[\nabla_{\theta_i} \left( \mathcal{R}_{\bar{c}}^{\pi_{\theta_i},\mu}\right)^m V_i(x)\right],
\end{align*}
for $m\geq 1$.Increasing $m$ has a similar effect as increasing $\bar{c}$ as both lengthen the effective lookahead horizon. Intriguingly, when we take $\mathcal{R}_{\bar{c}}^{\pi,\mu}$ to be the Q($\lambda$) operator with $\lambda=1$, the policy improvement objective $\left( \mathcal{R}_{\bar{c}}^{\pi_{\theta_i},\mu}\right)^m V_i(x)$ closely resembles the Taylor expansion policy optimization objective proposed in \citep{tang2020taylor}. A notable difference is that \citet{tang2020taylor} considered the special case where $V_i=V^\mu$ as the origin of the expansion, while here $V_i$ does not have to be the value function for any specific policy.

\section{Experiments}
\label{sec:exp}

We seek to answer the following questions: (Q1) Does multi-step improvement entail faster convergence to the optimal policy in tabular settings (Theorem~\ref{theorem:optimal})? (Q2) Does DoMo-AC introduce a bias-variance trade-off to estimating PG (Theorem~\ref{theorem:bias})? (Q3) Does DoMo-AC improve state-of-the-art policy based agents in large-scale settings?

\subsection{Tabular experiments} \label{sec:tabular}

To answer Q1, we start by empirically validating the speed-up of the convergence guarantee (predicted by Theorem~\ref{theorem:optimal}) entailed by DoMo-VI and DoMo-AC. We mainly compare three baselines: (1) one-step baseline VI (green), which consists of one-step policy improvement and evaluation $V_{i+1}=\mathcal{T}^{\pi_{i+1}}V_i$ where $\pi_{i+1}$ is one-step greedy; (2) multi-step policy evaluation (brown), which improves over VI with multi-step evaluation $V_{i+1}=\mathcal{R}_{\bar{c}}^{\pi_{i+1}}V_i$ for $\bar{c}=1$; (3) finally, the multi-step policy improvement algorithm where the policy $\pi_i = \pi_{\theta_i}$ is improved via $N$ gradient ascents with approximate gradient $\nabla_{\theta_i} \mathcal{R}_{\bar{c}}^{\pi_i,\mu}$ across all states. Formally, for $\forall 1\leq j\leq N$,
\begin{align*}
    \theta_{i+1}^{(j+1)} = \theta_{i+1}^{(j)} + \eta\frac{1}{|\mathcal{X}|}\sum_{x=1}^{|\mathcal{X}|}\nabla_{\theta_i^{(j)}} \mathcal{R}_{\bar{c}}^{\pi_{\theta_i^{(j)}},\mu}V(x_i),
\end{align*}
where we let $\theta_{i+1}=\theta_{i+1}^{(N)}$ as the final iterate of the gradient update. The value function is then updated via multi-step evaluation $V_{i+1}=\mathcal{R}_{\bar{c}}^{\pi_{i+1}}V_i$.
To study the impact of the degree of optimization, we consider $N\in\{1,10,100\}$ (purple, blue and red). By increasing $N$, the policy iterate $\pi_{\theta_{i+1}}$ gets closer to the optimal policy $\arg\max_{\pi_i}\mathcal{R}_{\bar{c}}^{\pi,\mu}V_i(x)$. All results are averaged over $100$ randomly generated MDPs. See Appendix~\ref{appendix:exp} for experimental details.

\begin{figure}[t]
    \centering
    \includegraphics[keepaspectratio,width=.4\textwidth]{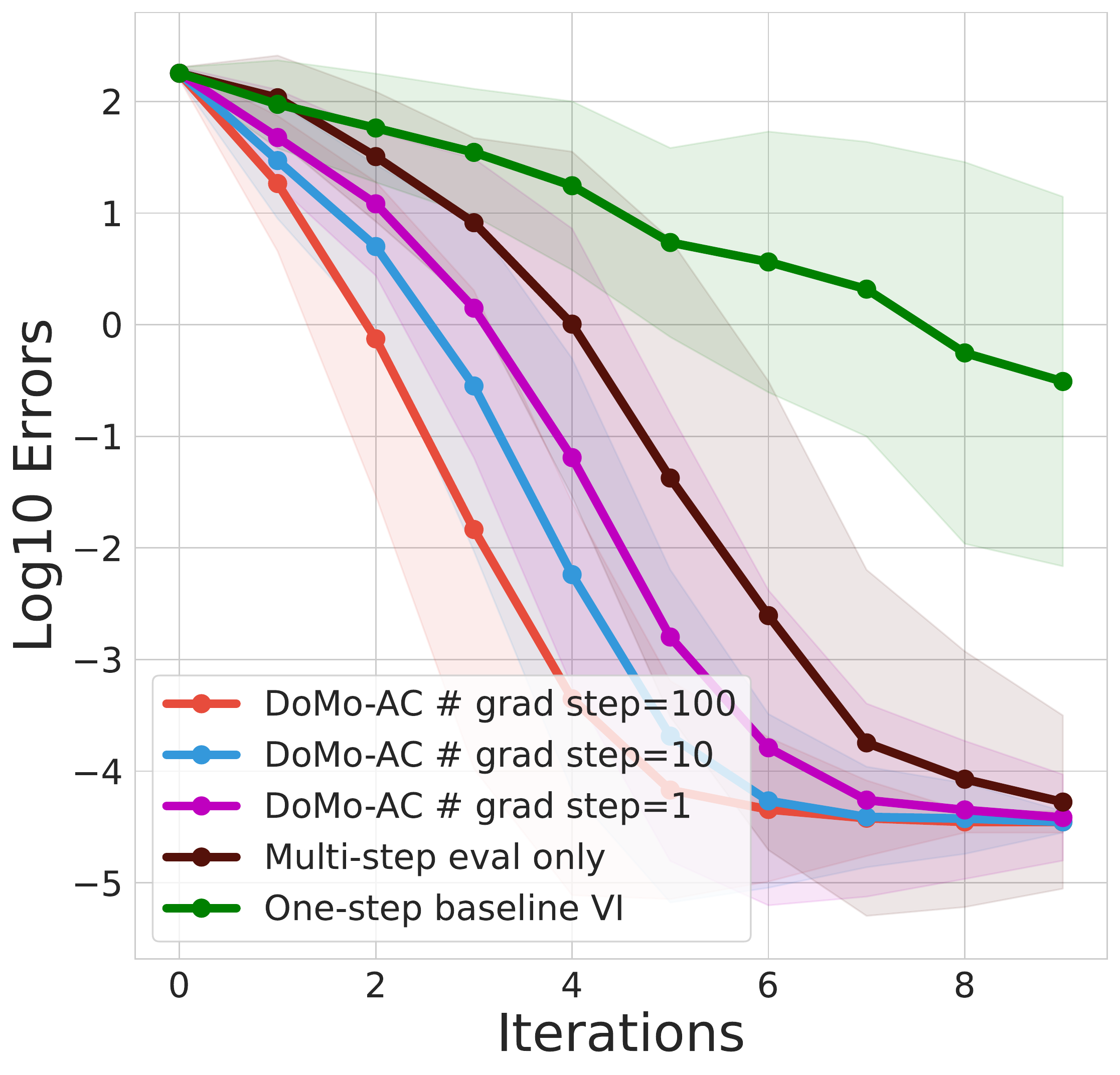}
    \caption{Evaluating the impact of approximate optimization of the policy improvement objective $\arg\max_{\pi \in \Pi} \mathcal{R}_{\bar{c}}^{\pi,\mu}V(x)$. The $y$-axis shows the value error $\left\lVert V^{\pi_i} - V^\ast \right\rVert$ on tabular MDPs. Throughout, we parameterize softmax policy and optimize the improvement objective with gradient ascent. Varying the number of gradient ascent steps, we see that as the number of steps increases, the improvement to convergence speed becomes more profound.}
    \label{fig:gradientstep}
\end{figure}

Figure~\ref{fig:gradientstep} shows the error $\left\lVert V^{\pi_i} -V^\ast \right\rVert_2$ as a function of iteration $i$. As expected, multi-step policy evaluation provides a major improvement over the VI baseline in accelerating the convergence. On top of that, as $N$ increases, multi-step policy improvement exhibits further performance improvements. This confirms the benefits of combining multi-step evaluation and improvement in the tabular settings where exact gradient computations are available. 

\paragraph{Stochastic gradient estimates in tabular settings.} 

To answer Q2, note that in DoMo-AC we use the stochastic update $\nabla_\theta \widehat{\mathcal{R}_{\bar{c}}^{\pi_\theta,\mu}}V(x)$ to update the policy parameter $\theta$. As discussed in Section~\ref{sec:ac}, the choice of $\bar{c}$ mediates a trade-off between bias and variance, on the approximation of $\nabla_\theta \widehat{\mathcal{R}_{\bar{c}}^{\pi_\theta,\mu}}V(x)$ to the true policy gradient $\nabla_\theta V^{\pi_\theta}(x)$. 

In Figure~\ref{fig:pg-gradient}, we examine such a bias-variance trade-off numerically. On a set of randomly generated MDPs, we calculate $\nabla_\theta \widehat{\mathcal{R}_{\bar{c}}^{\pi_\theta,\mu}}V(x)$ based on a fixed number of trajectories generated under behavior policy $\mu$. We then estimate the bias, variance and squared error of the policy gradient estimate against the ground truth $\nabla_\theta V^{\pi_\theta}(x)$. The results show that, as expected, when $\bar{c}$ increases from $0$ to $10$, the bias generally decreases, whereas the variance increases rapidly. This leads to an optimal middle ground (in this case $\log \bar{c}\approx 0$ and $\bar{c}\approx 1$) at which $\nabla_\theta \widehat{\mathcal{R}_{\bar{c}}^{\pi_\theta,\mu}}V(x)$ obtains the lowest squared error among this class of stochastic gradient estimates. Naturally, this trade-off will significantly impact the agent performance in large-scale settings, which we investigate next.

\begin{figure*}[t]
    \centering
    \includegraphics[keepaspectratio,width=.8\textwidth]{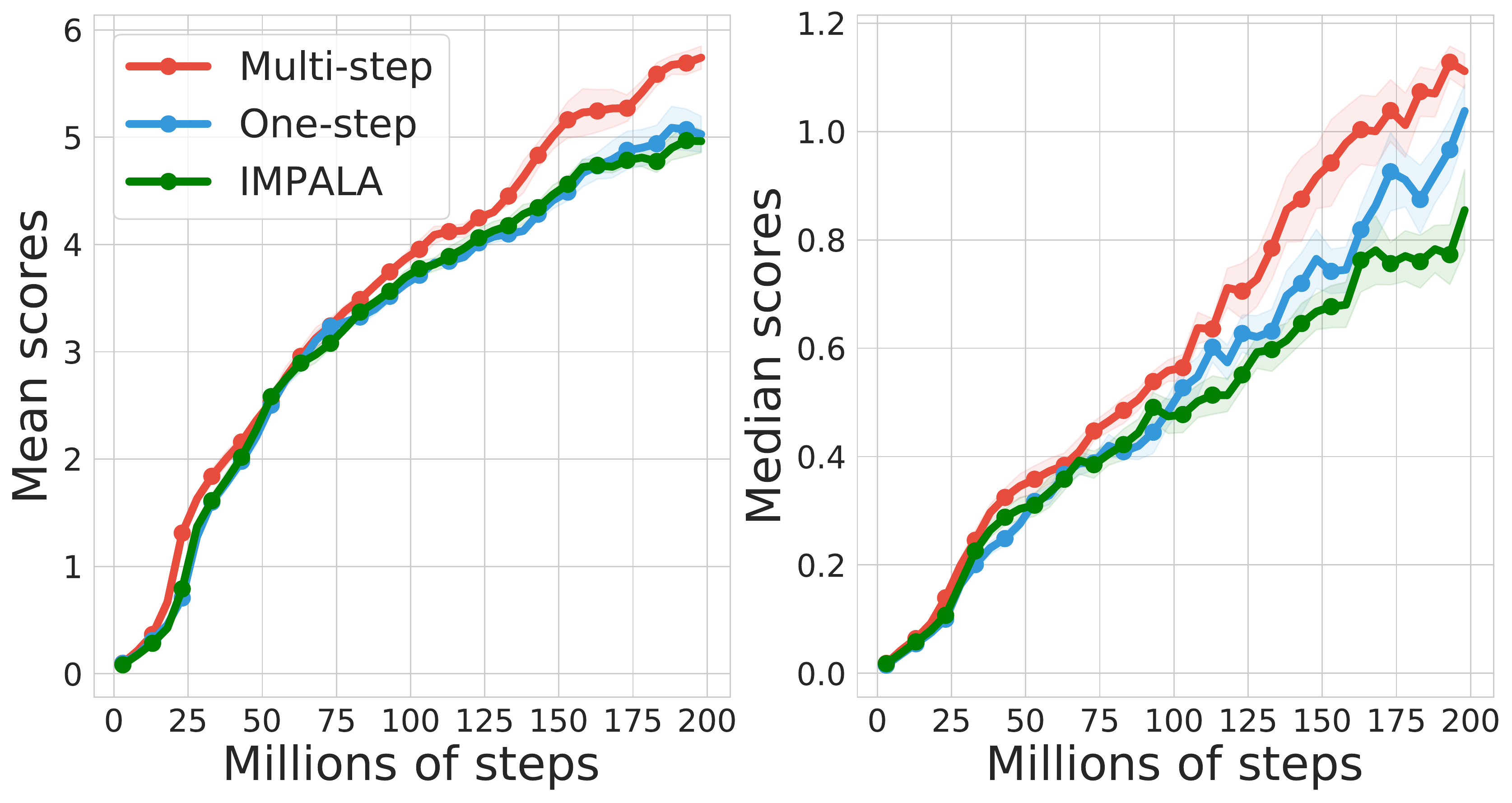}
    \caption{Comparing actor-critic algorithmic variants based on the IMPALA architecture \citep{espeholt2018impala}. We compare the DoMo-AC algorithm (Algorithm 1) instantiated with V-trace operator $\mathcal{R}_{\bar{c}}^{\pi,\mu}$ with $\bar{c}=0.5$; the one-step algorithm $\mathcal{R}_{\bar{c}}^{\pi,\mu}=\mathcal{T}^\pi$, which also be understood as the special case $\bar{c}=0$; and the IMPALA baseline. We report the evaluated  median and mean human normalized scores over $57$ Atari games, averaged across $5$ seeds. Overall, the DoMo-AC algorithm outperforms the one-step variant and the IMPALA baseline.}
    \label{fig:atari}
\end{figure*}

\subsection{Deep RL experiments}

To investigate the practical performance of DoMo-AC gradient update, we test different algorithmic variants with distributed actor-critic over architecture the Atari-57 games \citep{bellemare2013arcade}.

Our implementation is based on the IMPALA architecture \citep{espeholt2018impala}, an actor-critic algorithm with distributed actors and a centralized learner. The actors collect partial trajectories with the behavior policy $\mu$ and send to the learner with target policy $\pi_\theta$. Due to the latency of the actor-learner communication, the behavior policy uses a slightly stale copy of the policy parameter $\mu=\pi_{\theta_\text{old}}$, leading to inherent off-policyness during training $\pi_\theta\neq \mu$. By default, the learner maintains a policy network $\pi_\theta$ and a value network $V_\phi$.  Across all algorithmic variants we consider, the value networks are updated with the V-trace back-up targets \citep{espeholt2018impala} while we test different variants of actor updates. All algorithmic variants share hyper-parameters wherever possible.
See Appendix for further experiment details.

We compare a few algorithmic variants defined by different choices of the off-policy evaluation operators $\mathcal{R}_{\bar{c}}^{\pi_\theta,\mu}$
. For the multi-step variant, we choose V-trace with the trace coefficient threshold $\bar{c}$ as a tunable hyper-parameter. We find that $\bar{c}$ in between $0.3$ and $0.5$ works the best in practice and will report  the ablation results; for the one-step variant, we use the one-step operator $\mathcal{T}^\pi$, which can be understood as the special case $\bar{c}=0$. The baseline algorithm IMPALA \citep{espeholt2018impala} is closely related to the one-step variant, but with slightly different implementation details. We discuss such differences in Appendix~\ref{appendix:exp}.

In Figure~\ref{fig:atari}, we show the training performance curves of all algorithms. Each curve is an average over $5$ runs, with each run computed as either the mean or median human normalized scores across $57$ games. We find that the DoMo-AC implementation with V-trace $\bar{c}=0.5$ provides statistically significant improvements over one-step trace and IMPALA, implying the potential benefits of introducing multi-step gradient estimate.

\paragraph{Alternative off-policy evaluation operators.} Besides V-trace, other alternative trace coefficients such as tree-backup $c_t=\pi(A_t|X_t)$ \citep{precup2001off} and Q($\lambda$) $c_t=\lambda\in[0,1]$ \citep{harutyunyan2016q} all define valid off-policy evaluation operators \citep{munos2016safe}. We carry out a comparison with all such alternatives in Figure~\ref{fig:atari-all} in Appendix~\ref{appendix:exp}, where we show that V-trace obtains overall the best empirical performance.

\paragraph{Ablation on the trace coefficient threshold $\bar{c}$.} We next assess how sensitive the performance is to the trace coefficient threshold $\bar{c}$. We carry out experiments with $\bar{c}$ taking values in the range $[0,1]$ and graph the results in Figure~\ref{fig:atari-ablation} (Appendix~\ref{appendix:exp}). Going from $\bar{c}=0$ to $\bar{c}=1$, we find the best performance is obtained at the range $\bar{c}=0.3\sim 0.5$. The fact that $\bar{c}>0$ obtains the best performance demonstrates the practical utility of multi-step policy gradient estimate, compatible with the previous results. However, as $\bar{c}$ increases, the multi-step gradient estimate accumulates higher variance. Indeed, in the limit $\bar{c}\rightarrow\infty$, we have $c_t\rightarrow\rho_t$ and step-wise IS ratios can induce high variance to the overall estimates, which degrades the overall performance of the algorithm.

Intriguingly, here the optimal value of $\bar{c}\in [0.3,0.5]$ is noticeably lower than the typical value of the trace threshold applied in value-based learning (e.g., Retrace and V-trace all adopt $\bar{c}=1$ in their implementations by default \citep{munos2016safe,espeholt2018impala}). We speculate this might be because policy-based algorithms are generally more susceptible to high variance than value-based algorithms, and hence enjoy better performance when the estimates are of low variance.

\section{Conclusion}
We have proposed DoMo-VI, an extension of the classic VI algorithm which combines multi-step policy improvement with policy evaluation. Contrast to prior work, DoMo-VI enjoys theoretical speed-up to the optimal policy and is applicable in general off-policy settings. As a practical instantiation of the oracle algorithm, we propose DoMo-AC. DoMo-AC achieves the effect of multi-step improvement by applying a policy gradient estimator with a novel bias and variance trade-off. Compared to the baseline actor-critic algorithm, DoMo-AC generally enjoys more accurate approximation to the ground truth policy gradient. Implementing DoMo-AC with the IMPALA architecture, we observe a modest improvement from the baseline over the Atari game benchmarks. Possible future directions include adaptive methods for choosing the trace coefficient $\bar{c}$, and extensions of ideas of DoMo-VI more directly to value-based agents such as DQN.

\paragraph{Acknowledgements.} We thank anonymous reviewers for their valuable feedback to earlier drafts of this paper.

\bibliography{main}
\bibliographystyle{plainnat}

\clearpage
\onecolumn

\begin{appendix}

\section*{\centering APPENDICES: DoMo-AC: Doubly Multi-step Off-policy Actor-Critic Algorithm}

\section{Proof of theoretical results}

In this appendix, we provide missing proofs in the main paper. We begin with introducing some notations used in the proofs.

We denote an identity operator by $I$, which maps any real-valued function to itself.
Its domain will be clear from contexts.
For any Markov policy $\pi$, let $\Pi^{\pi}$ denote an operator that maps any bounded real-valued function $Q$ over $\mathcal{X} \times \mathcal{A}$ to a real-valued function $\Pi^{\pi} Q$ over $\mathcal{X}$ defined by
\begin{align*}
    \left( \Pi^{\pi} Q \right)(x)
    =
    \sum_{a \in \mathcal{A}} \pi (a|x) Q (x, a)
    \text{ at every } x \in \mathcal{X}\,.
\end{align*}
For a scalar $\bar c \in (0, \infty)$, and a behavior policy $\mu$, a similar operator $\Pi_{\bar c}^{\pi, \mu}$ maps $Q$ to a real-valued function $\Pi_{\bar c}^{\pi, \mu} Q$ over $\mathcal{X}$ defined by
\footnote{Recall we assume that a behavior policy has the full support: $\mu (a|x) > 0$ for all $(x, a) \in \mathcal{X} \times \mathcal{A}$.}
\begin{align*}
    \left( \Pi_{\bar c}^{\pi, \mu} Q \right)(x)
    =
    \sum_{a \in \mathcal{A}} \mu (a|x) \min \left\{ \bar c , \frac{\pi (a|x)}{\mu (a|x)} \right\} Q (x, a)
    \text{ at every } x \in \mathcal{X}\,.
\end{align*}
Abusing notations, let $P$ denote an operator that maps any bounded real-valued function $V$ over $\mathcal{X}$ to a real-valued function $P V$ over $\mathcal{X} \times \mathcal{A}$ defined by
\begin{align*}
    \left( P V \right)(x, a)
    =
    \sum_{y \in \mathcal{X}} P (y|x, a) V (y)
    \text{ at every } (x, a) \in \mathcal{X} \times \mathcal{A}
\end{align*}
Its conjugate with the $\Pi^\pi$ and $\Pi_{\bar c}^{\pi, \mu}$ operators are denoted by $P^{\pi} := \Pi^\pi P$ and $P^{c \mu \wedge \pi} := \Pi_{\bar c}^{\pi, \mu} P$, respectively.
With these operators, the V-trace operator can be rewritten as follows:
\begin{align*}
    \mathcal{R}_{\bar c}^{\pi, \mu} V
    =
    V + \left(I - \gamma P^{\bar c \mu \wedge \pi} \right)^{-1} \left( \Pi^{\pi} r + \gamma P^\pi V - V \right)
    =
    \left(I - \gamma P^{\bar c \mu \wedge \pi} \right)^{-1} \left( \Pi^{\pi} r + \gamma (P^\pi - P^{\bar c \mu \wedge \pi}) V \right),
\end{align*}
where $\left(I - \gamma P^{\bar c \mu \wedge \pi} \right)^{-1} := \sum_{t=0}^\infty \gamma^t \left( P^{\bar c \mu \wedge \pi} \right)^t$.
As the notation implies, it holds that $\left(I - \gamma P^{\bar c \mu \wedge \pi} \right) \left(I - \gamma P^{\bar c \mu \wedge \pi} \right)^{-1} = \left(I - \gamma P^{\bar c \mu \wedge \pi} \right)^{-1} \left(I - \gamma P^{\bar c \mu \wedge \pi} \right) = I$.

An operator, say $\mathcal{O}$, is said to be monotonic if $\mathcal{O} f \geq \mathcal{O} g$ for any pair of functions $f$ and $g$ such that $f \geq g$. All operators introduce above are monotonic.

\subsection{Proof of Lemma~\ref{lemma:stationary} (Optimal Markov Policy)}
\label{app:stationary policy proof}

Lemma~\ref{lemma:stationary} states that there exists a Markov policy $\pi$ such that
\begin{align*}
    \max_{p \in \Pi} \left( \mathcal{R}_{\bar c}^{p, \mu} V (x) \right)
    =
    \mathcal{R}_{\bar c}^{\pi, \mu} V (x)
    \text{ for all } x \in \mathcal{X},
\end{align*}
where $\Pi$ is the set of all Markov policies.
As $p$ on the left hand side may depend on $x \in \mathcal{X}$, the existence of $\pi$ is non-trivial.

For a fixed $V$ and $\mu$, let $\pi_x$ be a policy such that
$
    \pi_x := \arg\max_{p \in \Pi} \left( \mathcal{R}_{\bar c}^{p, \mu} V (x) \right).
$
Note that it is dependent on $x$, and there may be multiple policies that maximize the right hand side.
If it is not unique, pick up one arbitrarily.
Furthermore, let $\pi$ be a Markov policy such that
$
    \pi (\cdot | x)
    :=
    \pi_x (\cdot | x)
$
for all $x \in \mathcal{X}$.
By definition, for any Markov policy $\pi$ and any state $x \in \mathcal{X}$,
\begin{align*}
    &\mathcal{R}_{\bar c}^{\pi, \mu} V (x)
    \\
    &\leq
    \left(I - \gamma P^{c \mu \wedge \pi_x} \right)^{-1} \left( \Pi^{\pi_x} r + \gamma \left( P^{\pi_x} - P^{\bar c \mu \wedge \pi_x} \right) V \right) (x)
    \\
    &\leq
    \left( \Pi^{\pi_x} r + \gamma \left( P^{\pi_x} - P^{\bar c \mu \wedge \pi_x} \right) V \right) (x)
    +
    \gamma P^{c \mu \wedge \pi_x} \left(I - \gamma P^{c \mu \wedge \pi_x} \right)^{-1} \left( \Pi^{\pi_x} r + \gamma \left( P^{\pi_x} - P^{\bar c \mu \wedge \pi_x} \right) V \right) (x)
    \\
    &=
    \left( \Pi^{\pi} r + \gamma \left( P^{\pi} - P^{\bar c \mu \wedge \pi} \right) V \right) (x)
    +
    \gamma P^{c \mu \wedge \pi} \left(I - \gamma P^{c \mu \wedge \pi_x} \right)^{-1} \left( \Pi^{\pi_x} r + \gamma \left( P^{\pi_x} - P^{\bar c \mu \wedge \pi_x} \right) V \right) (x)\,,
\end{align*}
where the last line follows since $\pi (\cdot | x) = \pi_x (\cdot | x)$ by definition, and thus, $\Pi_c^{\pi_x, \mu} Q (x) = \Pi_c^{\pi, \mu} Q (x)$ for any bounded real-valued function $Q$ over $\mathcal{X} \times \mathcal{A}$.
Now, note that the second term is $\gamma P^{c \mu \wedge \pi} \mathcal{R}_{\bar c}^{\pi_x, \mu} V (x)$, and
\begin{align*}
    &P^{c \mu \wedge \pi} \mathcal{R}_{\bar c}^{\pi_x, \mu} (x)
    =
    \mathbb{E}_{y \sim P (\cdot | x, a), a \sim \mu (\cdot | x)}
    \left[
        \min \left\{ \bar c , \frac{\pi (a|x)}{\mu (a|x)} \right\} \mathcal{R}_{\bar c}^{\pi_x, \mu} V (y)
    \right]\,.
\end{align*}
Therefore, applying the same argument to $\mathcal{R}_{\bar c}^{\pi_x, \mu} V (y)$, we can conclude that
\begin{align*}
    \max_{\pi \in \Pi} \left( \mathcal{R}_{\bar c}^{\pi, \mu} V (x) \right)
    \leq
    \mathcal{R}_{\bar c}^{\pi, \mu} V (x).
\end{align*}

\subsection{Proof of Theorem~\ref{theorem:optimal} (Convergence Rate to Optimality)}

We upper-bound $\heartsuit$ and $\spadesuit$ in the following equation:
\begin{align*}
    V^* - V^{\pi_i}
    =
    \underbrace{V^* - \mathcal{R}_{\bar c}^{\pi_i, \mu} V_{i-1}}_{:= \heartsuit}
    + \underbrace{\mathcal{R}_{\bar c}^{\pi_i, \mu} V_{i-1} - V^{\pi_i}}_{:= \spadesuit}.
\end{align*}
For brevity, we let $\Pi^* := \Pi^{\pi^*}$, $P^* := \Pi^{\pi^*} P$, $\Pi^{*, \mu} := \Pi_{\bar c}^{\pi^*, \mu}$, $P^{*, \mu} := \Pi_{\bar c}^{\pi^*, \mu} P$, $\mathcal{R}^{*, \mu} := \mathcal{R}_{\bar c}^{\pi^*, \mu}$, $\Pi_j := \Pi_{\bar c}^{\pi_j, \mu}$, $P_j := \Pi_{\bar c}^{\pi_j, \mu} P$, and $\mathcal{R}_j := \mathcal{R}_{\bar c}^{\pi_j, \mu}$.

\paragraph{Upper-bound for $\heartsuit$.}
By definition, $\mathcal{R}_i V_{i-1} \geq \mathcal{R}^{*, \mu} V_{i-1}$, and $V^* = \mathcal{R}^{*, \mu} V^*$. Therefore,
\begin{align*}
    \heartsuit
    \leq
    \gamma \left(I - \gamma P^{*, \mu} \right)^{-1} \left( P^* - P^{*, \mu} \right) \left( V^* - V_{i-1} \right)
    =
    \gamma \left(I - \gamma P^{*, \mu} \right)^{-1} \left( P^* - P^{*, \mu} \right) \left( V^* - \mathcal{R}_{i-1} V_{i-2} \right).
\end{align*}
By induction on $i$,
$\heartsuit \leq (\Gamma^*)^i \left( V^* - V_0 \right)$,
where $\Gamma^* := \gamma \left(I - \gamma P^{*, \mu} \right)^{-1} \left( P^* - P^{*, \mu} \right)$.
As shown by \citet[around Eqn (12) in Appendix C]{munos2016safe}, $\Gamma^*$ is monotonic, and $\Gamma^* e \leq \eta^\ast e \leq \gamma e$, where $e$ is a constant function over $\mathcal{X}$ outputting $1$ everywhere.
Thus, $\heartsuit \leq (\eta^\ast)^i \left\| V^* - V_0 \right\|_\infty e$.
As both $V^*$ and $V_0$ are bounded by $1/(1-\gamma)$, $\left\| V^* - V_0 \right\|_\infty \leq 2 / (1-\gamma)$.

\paragraph{Upper-bound for $\spadesuit$.}
It holds that $V^{\pi_i} = \mathcal{R}_i V^{\pi_i}$. Therefore,
\begin{align*}
    \spadesuit
    &=
    \gamma \left(I - \gamma P_i \right)^{-1} \left( P^{\pi_i} - P_i \right) \left( V_{i-1} - V^{\pi_i} \right)
    \\
    &=
    \gamma \left(I - \gamma P_i \right)^{-1} \left( P^{\pi_i} - P_i \right) \left( V_{i-1} - \mathcal{R}_i V_{i-1} + \spadesuit \right)
    \\
    &=
    \gamma \left(I - \gamma P^{\pi_i} \right)^{-1} \left( P^{\pi_i} - P_i \right) \left( V_{i-1} - \mathcal{R}_i V_{i-1} \right)\,,
\end{align*}
where the last line follows since
\begin{align*}
    \spadesuit - \gamma \left(I - \gamma P_i \right)^{-1} \left( P^{\pi_i} - P_i \right) \spadesuit
    =
    \left(I - \gamma P_i \right)^{-1} \left( I - \gamma P_i - \gamma P^{\pi_i} + \gamma P_i \right) \spadesuit
    =
    \left(I - \gamma P_i \right)^{-1} \left( I - \gamma P^{\pi_i} \right) \spadesuit\,.
\end{align*}
By definition,
\begin{align*}
    V_{i-1} - \mathcal{R}_i V_{i-1}
    &=
    \mathcal{R}_{i-1} V_{i-2} - \mathcal{R}_i V_{i-1}
    \\
    &\leq
    \mathcal{R}_{i-1} V_{i-2} - \mathcal{R}_{i-1} V_{i-1}
    \\
    &=
    \gamma \left(I - \gamma P_{i-1} \right)^{-1} \left( P^{\pi_{i-1}} - P_{i-1} \right) \left( V_{i-2} - V_{i-1} \right)
    \\
    &=
    \gamma \left(I - \gamma P_{i-1} \right)^{-1} \left( P^{\pi_{i-1}} - P_{i-1} \right) \left( V_{i-2} - \mathcal{R}_{i-1} V_{i-2} \right)\,.
\end{align*}
By induction, we deduce that $V_{i-1} - \mathcal{R}_i V_{i-1} \leq \Gamma_{i-1} \cdots \Gamma_1 \left( V_0 - \mathcal{R}_1 V_0 \right)$, where $\Gamma_j := \gamma \left(I - \gamma P_j \right)^{-1} \left( P^{\pi_j} - P_j \right)$\,.
As
\begin{align*}
    V_0 - \mathcal{R}_1 V_0
    &=
    V_0 - V^{\pi_1} + \mathcal{R}_1 V^{\pi_1} - \mathcal{R}_1 V_0
    \\
    &=
    V_0 - V^{\pi_1} + \gamma \left(I - \gamma P_1 \right)^{-1} \left( P^{\pi_1} - P_1 \right) \left( V^{\pi_1} - V_0 \right)\,,
\end{align*}
we conclude that
\begin{align*}
    \spadesuit
    \leq
    \gamma \left(I - \gamma P^{\pi_i} \right)^{-1} \left( P^{\pi_i} - P_i \right) \Gamma_{i-1} \cdots \Gamma_1 \left( V_0 - V^{\pi_1} + \gamma \left(I - \gamma P_1 \right)^{-1} \left( P^{\pi_1} - P_0 \right) \left( V^{\pi_1} - V_0 \right) \right)\,.
\end{align*}
As shown by \citet{munos2016safe}, $P^{\pi_i} - P_i$ is monotonic, and $(P^{\pi_i} - P_i) e \leq e$, where $e$ is a constant function over $\mathcal{X}$ outputting $1$ everywhere.
Furthermore, $\Gamma_j$ is monotonic, and there exists some scalar $\kappa_j$ such that $\Gamma_j e \leq \kappa_j e \leq \gamma e$.
Thus,
\begin{align*}
    \spadesuit
    \leq
    \frac{\gamma \kappa_{i-1} \cdots \kappa_1}{1-\gamma} (1 + \kappa_1) \left\| V^{\pi_1} - V_0 \right\|_\infty e
    \leq
    \frac{\gamma \kappa_{i-1} \cdots \kappa_1}{1-\gamma} (1 + \gamma) \left\| V^{\pi_1} - V_0 \right\|_\infty e\,.
\end{align*}
Because both $V^{\pi_1}$ and $V_0$ are bounded by $1/(1-\gamma)$, $\left\| V^{\pi_1} - V_0 \right\|_\infty \leq 2 / (1-\gamma)$.

\paragraph{Combining Together.}
From those bounds and noting that $\gamma (1+\gamma) / (1-\gamma) + 1 \leq 2 / (1-\gamma)$, we conclude the proof.

\subsection{Proof of Theorem~\ref{theorem:bias}}

For notational simplicity, let $V_1\coloneqq \mathcal{R}_{\bar{c}}^{\pi_\theta,\mu}V$. In the below, we consider the gradient with respect to the $j$-th component of $\theta$. Then $\nabla_{\theta_j} V_1(x)$ is a vector of size $\mathbb{R}^{|\mathcal{X}|}$. Now, let $R^{\pi_\theta}\in\mathbb{R}^{|\mathcal{X}|}$ be the vector of reward such that $R^{\pi_\theta}(x)\coloneqq \sum_a r(x,a)\pi_\theta(a|x)$.
Plugging in the definition of the operator $\mathcal{R}_{\bar{c}}^{\pi_\theta,\mu}$ we have
\begin{align*}
    V_1 = \left(I-\gamma P^{c\mu} \right)^{-1} R^{\pi_\theta} +  \left(I-\gamma P^{c\mu} \right)^{-1}  \gamma\left(P^{\pi_\theta}-P^{c\mu}\right)V.
\end{align*}
Since $V^{\pi_\theta}$ is the fixed point of the operator, we can subtract both sides by $V^{\pi_\theta}$. This produces
\begin{align*}
     V_1 - V^{\pi_\theta} = \left(I-\gamma P^{c\mu} \right)^{-1}  \gamma\left(P^{\pi_\theta}-P^{c\mu}\right)\left(V - V^{\pi_\theta}\right).
\end{align*}

When the trace cofficient $c$ is smoothly differentiable in $\pi$, and under Assumption~\ref{assume:smooth}, we deduce that $\left(I-\gamma P^{c\mu} \right)^{-1}  \gamma\left(P^{\pi_\theta}-P^{c\mu}\right)$ is differentiable in $\theta_i$.
Let $g_1\coloneqq \nabla_{\theta_j}V_1\in \mathbb{R}^{|\mathcal{X}|}$ and $g \coloneqq \nabla_{\theta_j}V^{\pi_\theta}\in \mathbb{R}^{|\mathcal{X}|}$. The gradient vector $g_1$ satisfies the following recursive equation, with $g_0\coloneqq \nabla_{\theta_j}V=0$ obtained by taking derivative of both sides above w.r.t. $\theta_i$,
\begin{align*}
    g_1-g=\nabla_{\theta_i}\left[\left(I-\gamma P^{c\mu} \right)^{-1}  \gamma\left(P^{\pi_\theta}-P^{c\mu}\right)\right]\left(V-V^{\pi_\theta}\right)+\left(I-\gamma P^{c\mu} \right)^{-1}  \gamma\left(P^{\pi_\theta}-P^{c\mu}\right)\left(g_0-g\right).
\end{align*}
When $V=V^{\pi_\theta}$, the first term vanishes and note that the matrix $ \left(I-\gamma P^{c\mu} \right)^{-1}  \gamma\left(P^{\pi_\theta}-P^{c\mu}\right) $ has operator norm upper bounded by $\eta\leq \gamma$ \citep{munos2016safe}. We hence deduce the following inequality which concludes the proof
\begin{align*}
\left\lVert g_1-g\right\rVert_\infty\leq \eta\left\lVert g_0\right\rVert_\infty.   
\end{align*}

\subsection{Proof of Theorem~\ref{theorem:unbiased}}

By construction of the V-trace operator, it is straightforward to verify that the following
\begin{align*}
    \widehat{\mathcal{R}_{\bar{c}}^{\pi_{\theta},\mu}}V(x) \coloneqq V(x) + \sum_{t=0}^\infty \gamma^t c_{0:t-1} \rho_t\delta_t
\end{align*}
is an unbiased estimate to $\mathcal{R}_{\bar{c}}^{\pi_\theta,\mu}V(x)$. Now, since we assume the trajectory is of finite length almost surely and since the importance sampling ratio $\rho_t\leq \max_{x,a}\frac{\pi_\theta(a|x)}{\mu(a|x)}$ is upper bounded, we can verify that we can apply the dominated convergence theorem to the limiting sequence 
\begin{align*}
\frac{1}{\delta_j}\left(\widehat{\mathcal{R}_{\bar{c}}^{\pi_{\theta+\delta_j},\mu}}V(x) - \widehat{\mathcal{R}_{\bar{c}}^{\pi_{\theta+\delta},\mu}}V(x)\right)
\end{align*} 
with $\left\lVert \delta_j\right\rVert_2\rightarrow 0$, which implies $
    \mathbb{E}_\mu\left[\nabla_\theta \widehat{\mathcal{R}_{\bar{c}}^{\pi_{\theta},\mu}}V(x)\right] = \nabla_\theta \mathcal{R}_{\bar{c}}^{\pi_\theta,\mu}V(x).
$ and hence $\nabla_\theta \widehat{\mathcal{R}_{\bar{c}}^{\pi_{\theta},\mu}}V(x)$ is an unbiased gradient estimate.

\section{Experiment details and additional results}
\label{appendix:exp}

We present further experiment details and results.
\subsection{Tabular experiments on VI}

\paragraph{Figure~\ref{fig:rate}.} We compare DoMo-VI, multi-step policy evaluation, multi-step policy optimization  and one-step baseline VI. All experiments are carried out on tabular MDPs with $|\mathcal{X}|=20$ states $|\mathcal{A}|=5$ actions. The transition $p(\cdot|x,a)$ is generated as Dirichlet random variable with parameter $(\alpha,...\alpha)\in\mathbb{R}^\mathcal{X}$ for $\alpha=0.01$. The reward $R_0$ is sampled from a standard normal distribution and kept fixed. The discount factor $\gamma=0.9$. For all multi-step variants, we set $\bar{c}=10$. 

We carry out recursions based on different algorithms and report the approximation error to the optimal value function $\left\lVert V^{\pi_i} - V^\ast\right\rVert_2$. All results are repeated $100$ times with randomly generated MDPs. For implementing DoMo-VI and multi-step policy optimization, we need to approximately solve the optimization problem
$
    \arg\max_{\pi \in \Pi} \mathcal{R}_{\bar{c}}^{\pi,\mu}V(x).
$.
To this end, we parameterize policy $\pi_\theta(a|x)=\text{softmax}(\theta(x,a))$ and carry out gradient ascent on the objective below until convergence.
\begin{align}
    L(\theta) = \frac{1}{|\mathcal{X}|}\sum_{x=1}^{|\mathcal{X}|} \mathcal{R}_{\bar{c}}^{\pi_\theta,\mu}V(x).\label{eq:uniform-policy-objective}
\end{align}

\paragraph{ Figure~\ref{fig:gradientstep}.} We compare DoMo-AC, multi-step policy evaluation and one-step baseline VI. All experiments are carried out using the same setup as above. Notably, DoMo-AC is an approximation to DoMo-VI in that the policy optimization stage is not necessarily carried out in full. At iteration $i$, let $\pi_g$ be the current greedy policy with respect to $V_i$, we initialize a softmax policy with parameter $\theta_{i+1}^{(1)}$ such that
\begin{align*}
    \theta_{i+1}^{(1)}(x,a) = \log \left(\pi_g(a|x) + 10^{-5}\right).
\end{align*}
This is such that the softmax policy defined with $\theta_{i+1}^{(1)}$ is close to $\pi_i$. This initialization is intended such that when there is no gradient update, the performance of DoMo-AC is similar to the multi-step policy evaluation baseline (with one-step greedy). We then carry out gradient updates on the objective $L(\theta_{i+1}^{(j)})$ as defined in Eqn~\eqref{eq:uniform-policy-objective} for $N$ steps. The final iterate $\theta_{i+1}^{(N)}$ is used for defining the policy $\pi_{i+1}$ at the next iteration. All results are repeated for $100$ times across randomly generated MDPs.

\subsection{Deep RL experiments}

All evaluation environments are the entire suite of Atari games \citep{bellemare2013arcade} consisting of $57$ levels. Since each level has a very different reward scale and difficulty, we report human-normalized scores for each level, calculated as $z_i = (r_i - o_i) / (h_i - o_i)$, where $h_i, o_i$ are performances of human and a random policy on level $i$ respectively. 

For all experiments, we report summarizing statistics of the human-normalized scores across all levels. For example, at any point in training, the mean human-normalized score is the mean statistics across $z_i,1\leq i\leq 57$.

\paragraph{Distributed training.} 
Distributed algorithms have led to significant performance gains on challenging domains \citep{nair2015massively,mnih2016,babaeizadeh2016reinforcement,barth2018distributed,horgan2018distributed}. Here, our focus is on recent state-of-the-art algorithms. In general, distributed agents consist of one central learner, multiple actors and optionally a replay buffer. The central learner maintains a parameter copy $\theta$ and updates parameters based on sampled data. Multiple actors each maintaining a slighted delayed parameter copy $\theta_{\text{old}}$ and interact with the environment to generate partial trajectories. Actors sync parameters from the learner periodically. In the actor-critic setting, the behavior policy is executed using the delayed copy such that $\mu=\pi_{\theta_\text{old}}$.

\paragraph{Details on the distributed architecture.} The general policy-based distributed agent follows the architecture design of IMPALA \citep{espeholt2018impala}, i.e. a central GPU learner and $N=512$ distributed CPU actors. The actors keep generating data by executing their local copies of the policy $\mu$, and sending data to the queue maintained by the learner. The parameters are periodically synchronized between the actors and the learner, as discussed above.

The architecture details are the same as those in \citep{espeholt2018impala}. For completeness, we present some important details below, please refer to the original paper for other missing details. See the paper for further details.

The policy/value function networks are both trained by RMSProp optimizers \citep{tieleman2012lecture} with learning rate $\alpha=5\cdot10^{-4}$ and no momentum. To encourage exploration, the policy loss is augmented by an entropy regularization term with coefficient $c_e=0.01$ and  baseline loss with coefficient $c_v=0.5$, i.e. the full loss $L = L_{\text{policy}} + c_vL_\text{value} + c_e L_{\text{entropy}}$. These single hyper-parameters are selected according to Appendix D of \citep{espeholt2018impala}.

Actors send partial trajectories of length $T=20$ to the learner.
For robustness of the training, rewards $R_t$ are clipped between $[-1,1]$. We adopt frame stacking and sticky actions as commonly practiced  \citep{mnih2013playing}. The discount factor $\gamma=0.99$ for calculating the baseline estimations.

\paragraph{V-trace value learning implementations.} The targets for value learning $V_\text{target}(X_t)$ in Algorithm 1 are computed via V-trace. V-trace is a competitive baseline for correcting off-policy data \citep{espeholt2018impala}. Given a partial trajectory $(X_t,A_t,R_t)_{t=1}^T$, let $\tilde{\rho}_t = \min\{\bar{\rho},\rho_t\}$ be the truncated IS ratio. Let $v(x)$ be the a certain  value function baseline (e.g., we let the baseline be computed by the value network $v(x)=V_\phi(x)$). V-trace targets are calculated recursively for all $1\leq t\leq T$ backward in time:
\begin{align}
    V_\text{target}(X_t) = v(X_t) + \tilde{\rho}_t \delta_t + \gamma c_t\left(V_\text{target}(X_{t+1}) - v(X_t)\right), \label{eq:V-trace-recursive}
\end{align}
where $\tilde{\rho}_t=\min(\bar{\rho},\rho_t)$ is a truncated IS ratio and $c_t=\min(\bar{c},\rho_t)$ is the trace coefficient. When $t=T$, we initialize $V_\text{target}(X_t)=v(X_t)$. In practice, it is common to set $\bar{\rho}<\infty$ to avoid explosion of the IS ratio; though this introduces extra bias into the gradient estimate. The value function baseline is then trained to approximate these targets $V_\phi(x) \approx v(x)$. Following \citep{espeholt2018impala}, we set $\bar{\rho}=\bar{c}=1$. 

\paragraph{Implementation details of DoMo-AC.} We build the DoMo-AC gradient estimate on top of the V-trace recursive estimate in Eqn~\eqref{eq:V-trace-recursive}. Note that we can think of $V_\text{target}(X_t)$, as computed above, as a function of parameter $\theta$ as $c_t=\min(\rho_t,\bar{c})$ where $\rho_t=\pi_\theta(A_t|X_t)/\mu(A_t|X_t)$. We can understand $V_\text{target}(X_t)$ as effectively the  estimated back-up target $\widehat{\mathcal{R}_{\bar{c}}^{\pi_\theta,\mu}}v(X_t)$ and compute the DoMo-AC gradient estimate by differentiating through $V_\text{target}(X_t)$ via auto-diff. In calculating the back-up targets for value learning, we use $v(X_t)=V_\phi(X_t)$; however, for estimating policy gradient, we find that the algorithm works better with $v(X_t)=V_\text{target}(X_t)$. We speculate that this is because policy gradient estimates would benefit from a more accurate baseline, and the V-trace estimate $V_\text{target}(X_t)$ provides a more accurate approximation to the true value function compared to the baseline.

\paragraph{Alternative evaluation operators for deep RL experiments.} All operators take the same form as the V-trace operator in Eqn~\eqref{eq:V-trace} but differ in the choice of trace coefficient $c_t$. We consider a few alternatives: (1) By default, the V-trace operator with Retrace trace $c_t=\min(\rho_t,\bar{c})$ with $\bar{c}=0.5$. We will examine the sensitivity to the threshold $\bar{c}$ in ablation study; (2) The one-step trace, $c_t=0$, which instantiates the actor-critic instantiation of the multi-step policy evaluation recursion. It turns out that such an algorithm closely resembles the original IMPALA implementation; (3) Tree back-up trace $c_t=\pi(A_t|X_t)$; (4) Q($\lambda$) trace with $c_t=\lambda=0.7$. Finally, we also compare with the IMPALA baseline \citep{espeholt2018impala}.

\begin{figure*}[t]
    \centering
    \includegraphics[keepaspectratio,width=.8\textwidth]{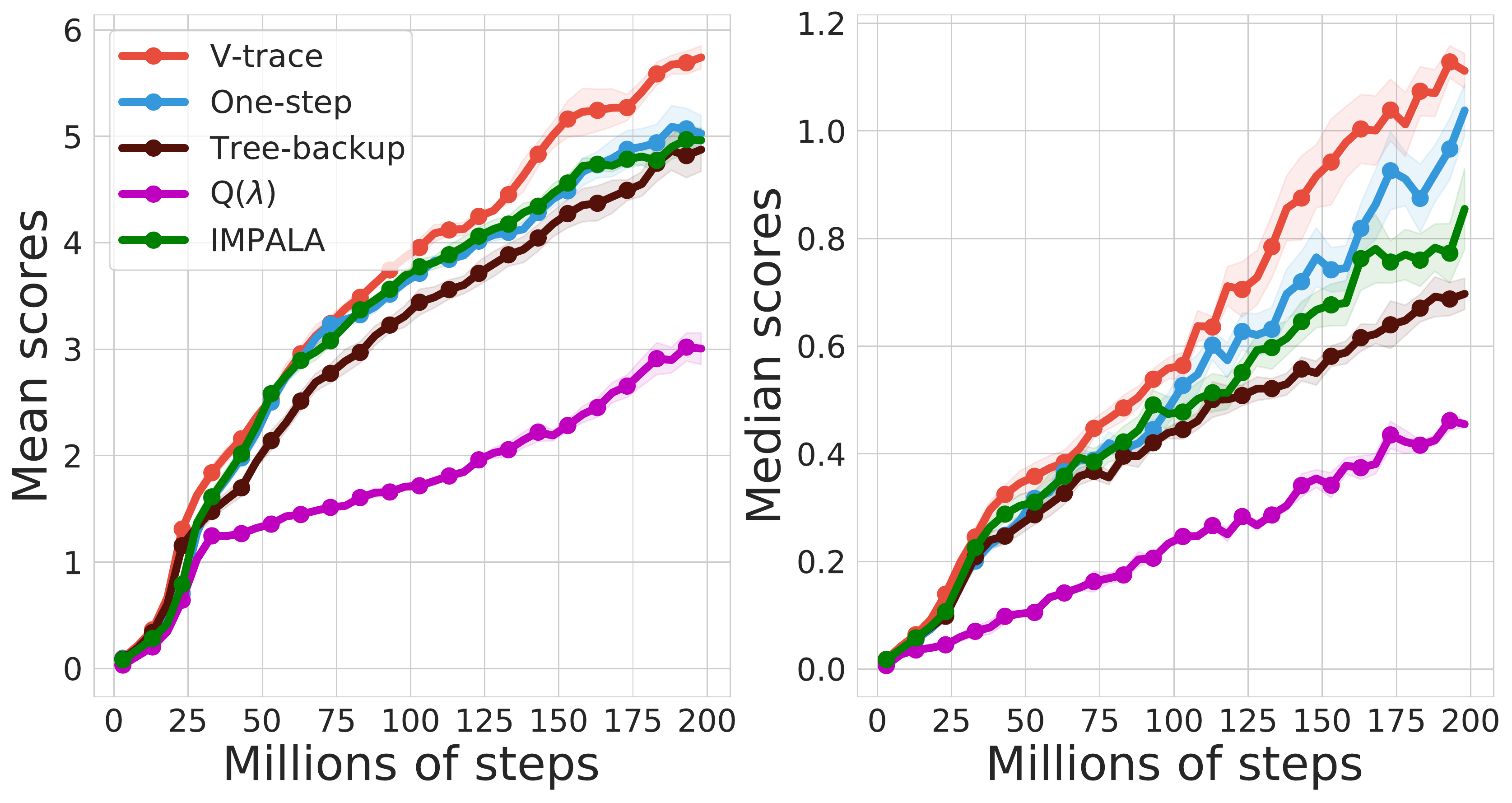}
    \caption{Full results for the Atari game suites, and comparison across various baseline operators. We show the mean and median performance of baseline algorithms across all $57$ Atari games. Overall, we see that V-trace retains performance advantage compared to other alternative off-policy evaluation operators when applied under the DoMo-AC framework.}
    \label{fig:atari-all}
\end{figure*}

\begin{figure}[t]
    \centering
    \includegraphics[keepaspectratio,width=.4\textwidth]{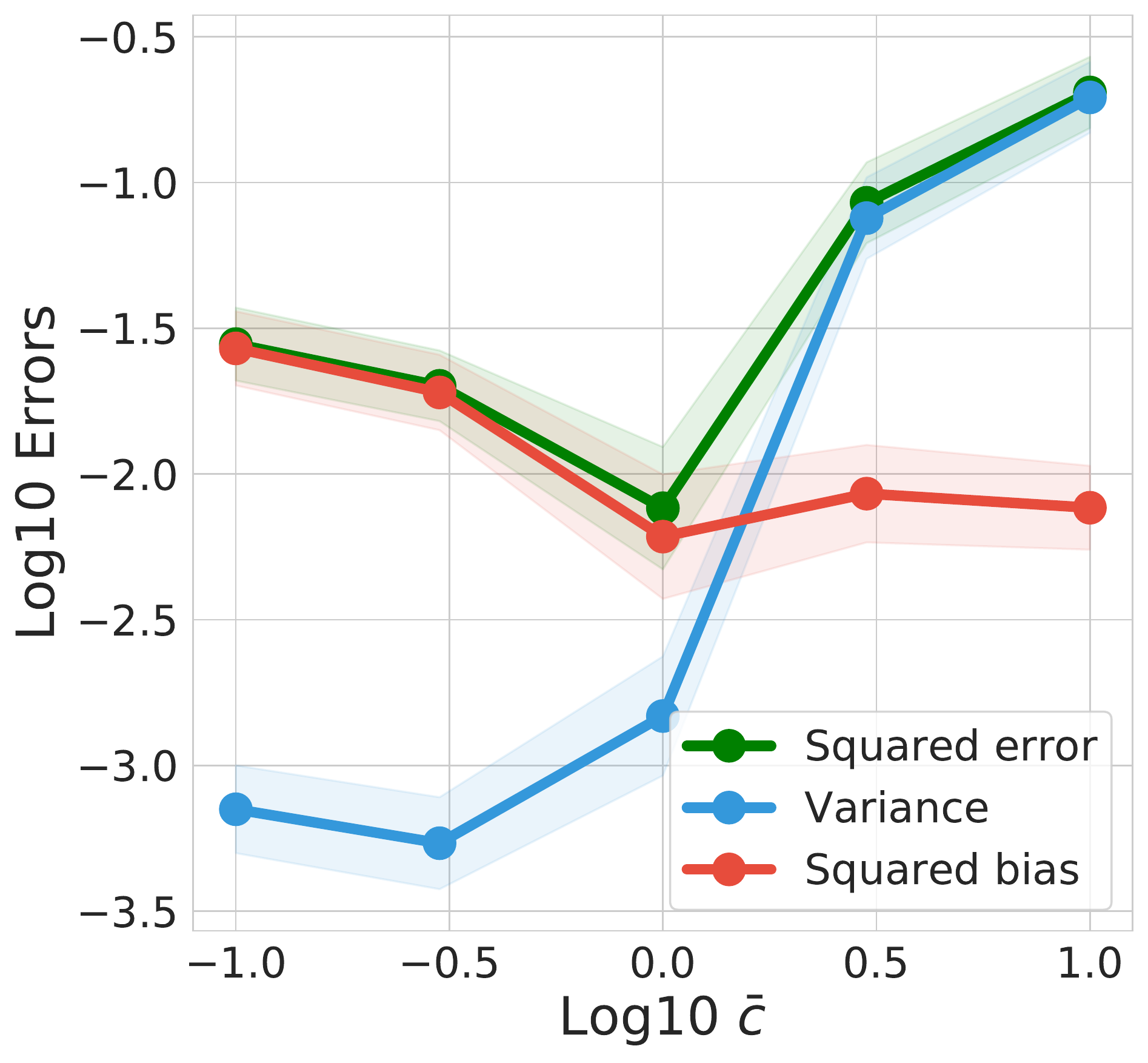}
    \caption{The bias-variance trade-off of the stochastic estimate $\nabla_\theta \widehat{\mathcal{R}_{\bar{c}}^{\pi_\theta,\mu}}V(x)$ against the true policy gradient $\nabla_\theta V^{\pi_\theta}(x)$ on a number of randomly generated MDPs. As $\bar{c}$ increases, the bias generally decreases but the variance increases. Overall, this leads to an optimal middle ground for the choice of $\bar{c}$. See Appendix~\ref{appendix:exp} for more details on the experimental setups.}
    \label{fig:pg-gradient}
\end{figure}

 \begin{figure}[t]
    \centering
    \includegraphics[keepaspectratio,width=.4\textwidth]{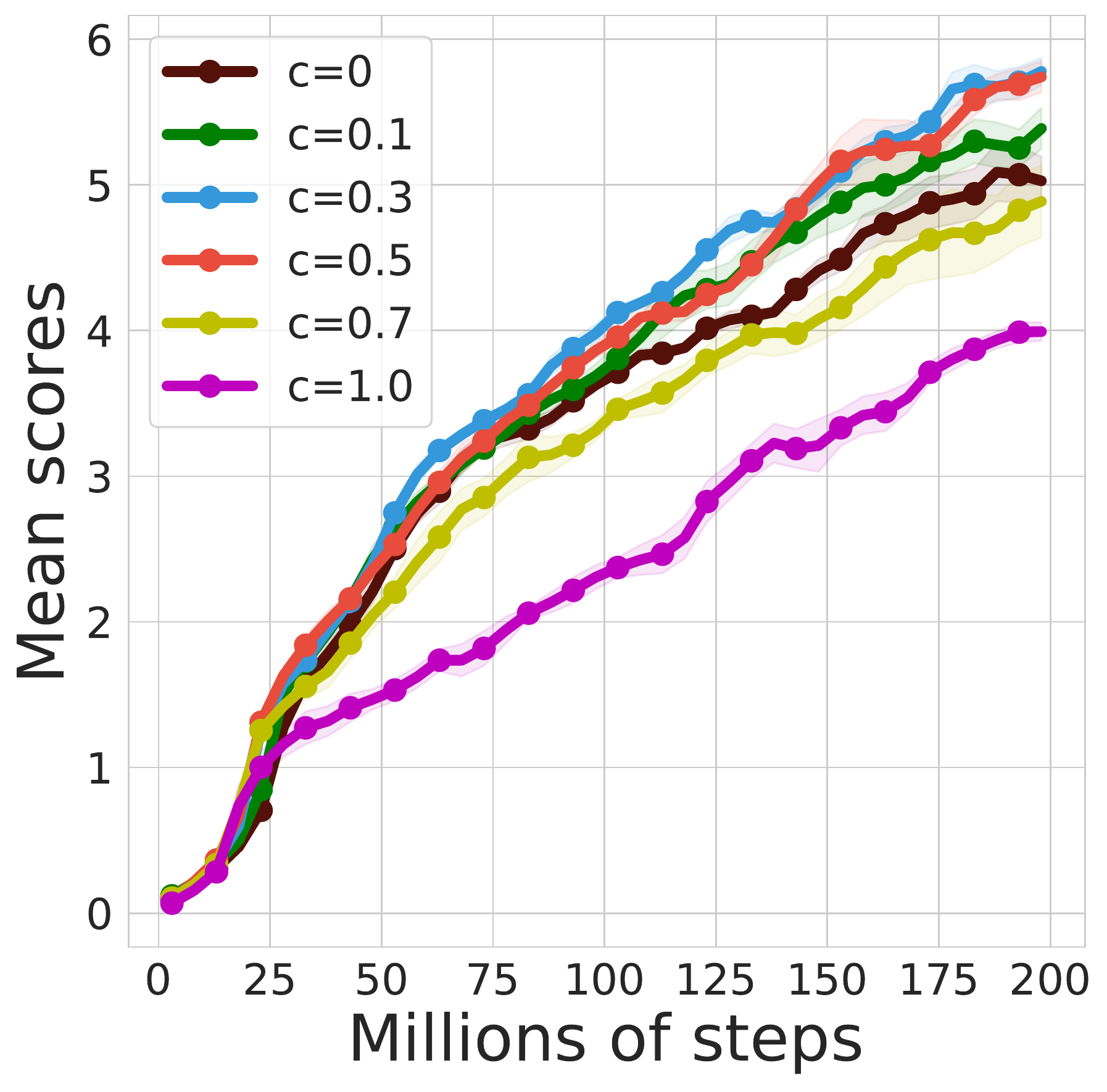}
    \caption{Ablation study on the effect of the trace coefficient threshold $\bar{c}$ for the V-trace operator in DoMo-AC algorithm. Going from $\bar{c}=0$ to $\bar{c}=1$, the evaluated performance throughout training first increases and then decreases. The best-performing value of $\bar{c}$ seems to be between $0.3$ and $0.5$, where the best bias-variance trade-off is obtained.}
    \label{fig:atari-ablation}
\end{figure}

\section{Discussion on truncated operators}

In tabular experiments, though the back-up target  $\nabla_\theta \mathcal{R}_{\bar{c}}^{\pi_\theta,\mu}V(x)$ is defined with an infinite horizon, it can be computed analytically using matrix inverse and auto-diff.
In large-scale experiments, gradients are computed based on sampled trajectories. Since the partial trajectories are of length $T$, we can understand the practical algorithm as being derived from the equivalent off-policy evaluation operator takes the truncated form  
\begin{align}
     \mathcal{R}_{T,\bar{c}}^{\pi,\mu}V(x) \coloneqq V(x) + \mathbb{E}_\mu\left[\sum_{t=0}^{T-1} \gamma^t c_{0:t-1} \rho_t\delta_t \right].
    \label{eq:V-trace-truncated}
\end{align}
The truncated operator enjoys similar theoretical properties as the non-truncated operator $ \mathcal{R}_{\bar{c}}^{\pi,\mu}$, such as the fixed point $V^\pi$ and accelerated contraction rate compared to the one-step operator.

\end{appendix}

\end{document}